\documentclass[10pt,twocolumn,letterpaper]{article}

\usepackage{cvpr}      

\usepackage{graphicx}
\usepackage{amsmath}
\usepackage{amssymb}
\usepackage{booktabs}
\usepackage{multirow}
\usepackage{xcolor}         
\usepackage{colortbl}
\usepackage{threeparttable}
\usepackage{amsmath}
\usepackage{float}
\usepackage[ruled,linesnumbered]{algorithm2e}

\usepackage[pagebackref,breaklinks,colorlinks]{hyperref}

\usepackage[capitalize]{cleveref}
\crefname{section}{Sec.}{Secs.}
\Crefname{section}{Section}{Sections}
\Crefname{table}{Table}{Tables}
\crefname{table}{Tab.}{Tabs.}

\def\cvprPaperID{526} 
\def\confName{CVPR}
\def\confYear{2023}

\begin{document}

\title{Geometric Visual Similarity Learning in 3D Medical Image Self-supervised Pre-training}

\author{Yuting He$^{1}$, Guanyu Yang$^{1}$\thanks{Corresponding author: yang.list@seu.edu.cn}, Rongjun Ge$^{2}$, Yang Chen$^{1}$, Jean-Louis Coatrieux$^{3}$, Boyu Wang$^{4}$, Shuo Li$^{5}$\\
$^{1}$Southeast University $^{2}$Nanjing University of Aeronautics and Astronautics \\ $^{3}$University of Rennes 1 $^{4}$Western University
$^{5}$Case Western Reserve University
}
\maketitle
\begin{abstract}
Learning inter-image similarity is crucial for 3D medical images self-supervised pre-training, due to their sharing of numerous same semantic regions. However, the lack of the semantic prior in metrics and the semantic-independent variation in 3D medical images make it challenging to get a reliable measurement for the inter-image similarity, hindering the learning of consistent representation for same semantics. We investigate the challenging problem of this task, i.e., learning a consistent representation between images for a clustering effect of same semantic features. We propose a novel visual similarity learning paradigm, Geometric Visual Similarity Learning, which embeds the prior of topological invariance into the measurement of the inter-image similarity for consistent representation of semantic regions. To drive this paradigm, we further construct a novel geometric matching head, the Z-matching head, to collaboratively learn the global and local similarity of semantic regions, guiding the efficient representation learning for different scale-level inter-image semantic features. Our experiments demonstrate that the pre-training with our learning of inter-image similarity yields more powerful inner-scene, inter-scene, and global-local transferring ability on four challenging 3D medical image tasks. Our codes and pre-trained models will be publicly available\footnote{\url{https://github.com/YutingHe-list/GVSL}}.
\end{abstract}
\section{Introduction}
\begin{figure}
\centering
\includegraphics[width=\linewidth]{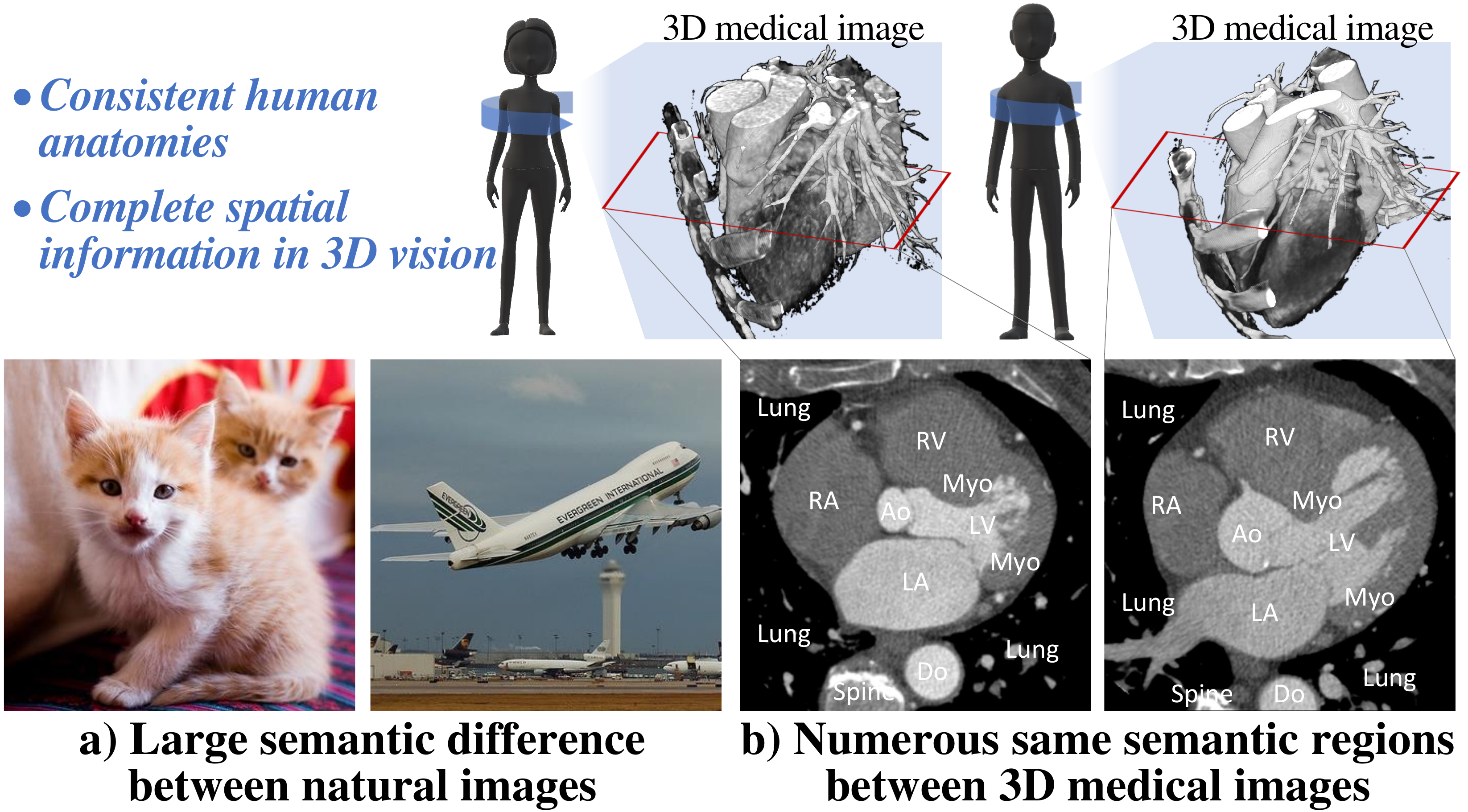}
\caption{Learning inter-image similarity is crucial for 3D medical image SSP. a) Natural images have large semantic difference between images whose inter-image similarity is weak. b) 3D medical images share numerous same semantic regions between images due to the consistent human anatomies and the complete spatial information in 3D vision, having large inter-image similarity.}
\label{fig:compare}
\end{figure}
Learning inter-image similarity \cite{milbich2021visual,zhang2022attributable,roth2020pads,you2022mine} is crucial for 3D medical image (e.g., CT, MR) self-supervised pre-training (SSP) \cite{jing2020self}. As shown in Fig.\ref{fig:compare}, different from natural images which are widely researched in SSP, 3D medical images share numerous same semantic regions due to the consistency of human anatomies \cite{netter2014atlas} and the complete spatial information in 3D vision \cite{sabharwal2015completeness}, bringing a strong prior for effective SSP. Therefore, it targets on constraining the pre-training network for a consistent representation of these same semantic regions between images without annotations. Once successful, it will bring great clustering effect for same semantic features, powerful representability of pre-trained network, and effective transferring for potential downstream tasks.

\begin{figure}
\centering
\includegraphics[width=\linewidth]{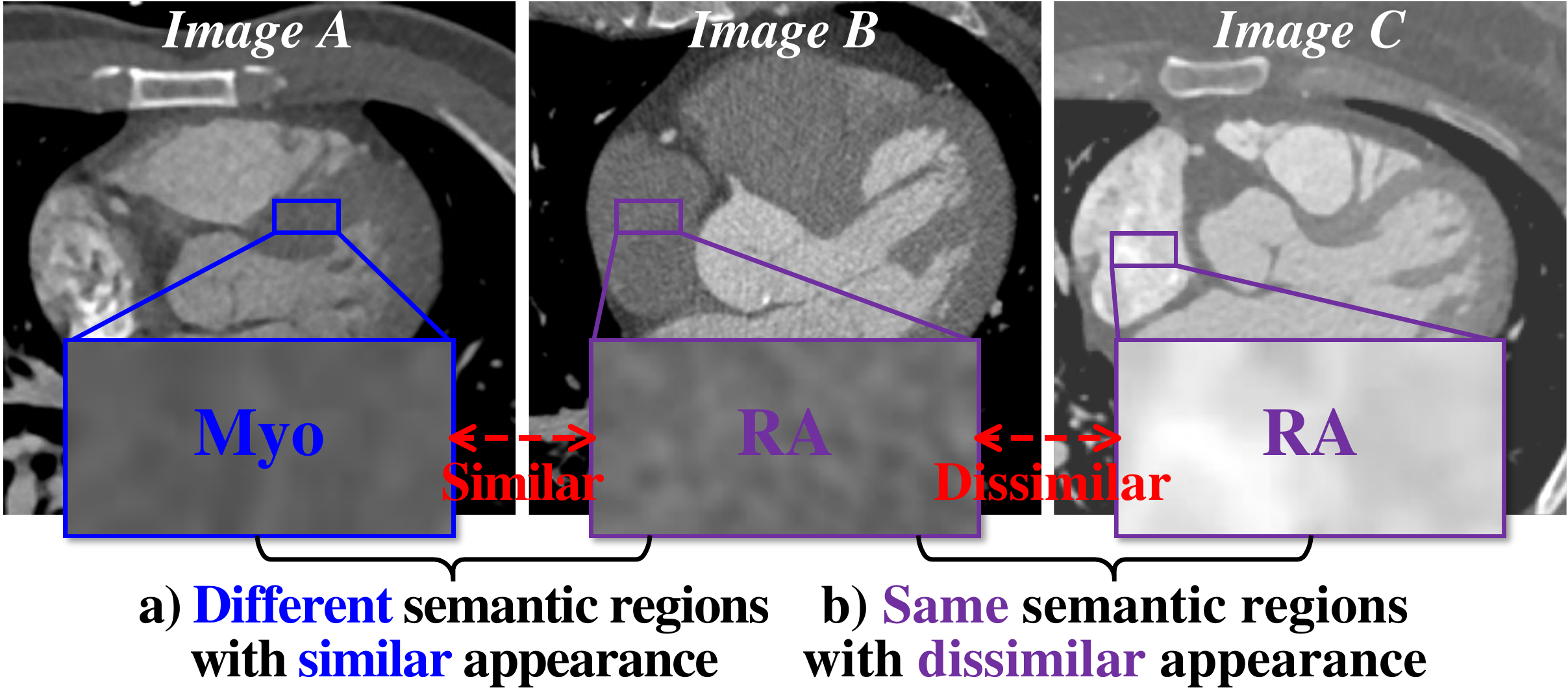}
\caption{It is challenging to measure a reliable inter-image similarity. a) There is a large similarity between the Myo and the RA regions between images A and B. b) Due to the variation of the scanning protocol, RA regions are different in images B and C.}
\label{fig:limitation}
\end{figure}

Although the existing SSP works have achieved promising results in their tasks, they are limited in the learning of inter-image similarity in 3D medical images. 1) Clustering-based SSP methods \cite{caron2018deep,li2021dense} measure the features' similarity between images for their clustering pattern in an embedding space, and learn to aggregate same cluster's features. However, they simply employ the Mahalanobis or Euclidean distance as the measurement function which is extremely interfered by images' semantics-independent variations (Fig.\ref{fig:limitation}). 2) Contrastive learning works \cite{Chen2021CVPR,chen2020simple} directly learn to separate their features for inter-image \emph{dissimilarity}. This violates the learning inter-image \emph{similarity} which is crucial in 3D images and will make the network represent distinct features for same semantic regions. Although some other contrastive learning works \cite{grill2020bootstrap,Chen2021CVPR,wang2022exploring} have removed the separation learning, they are still unable to learn the consistency of inter-image same semantics. 3) Generation-based methods \cite{liu2021self,komodakis2018unsupervised,vincent2010stacked,zhou2019models} construct pretext labels via designed transformation methods (e.g., rotation \cite{komodakis2018unsupervised}) and train networks to predict these labels. These methods implicitly impose a bias into SSP via manually designing the transformation methods. However, the bias extremely relies on the manual design which makes pre-training networks focus on the biased features of pretext labels and become sensitive to the change of scenario \cite{liu2021self}. 

Thinking the limitations in above existing works, the \emph{large-scale mis-measurement} for inter-image similarity is the key challenge in 3D medical SSP, interfering the discovery of semantics' correspondence and hindering the learning of consistent representation for same semantic regions. Semantic-independent variations (Fig.\ref{fig:limitation}) make the 3D medical images have different appearance. Different semantic regions have similar appearances and same semantic regions have different appearances between images. The direct measurement in the embedding space, like the clustering-based SSP methods \cite{caron2018deep,li2021dense}, is sensitive due to lack of semantic prior in their metrics. Therefore, in the non-supervision situation, once the features changed caused by the variations, these metrics will make mis-measurement of similarities for large-scale semantics, bringing their mis-correspondence. It will train network to aggregate the features with different semantic but similar appearance, causing mis-representation.

\begin{figure}
\centering
\includegraphics[width=\linewidth]{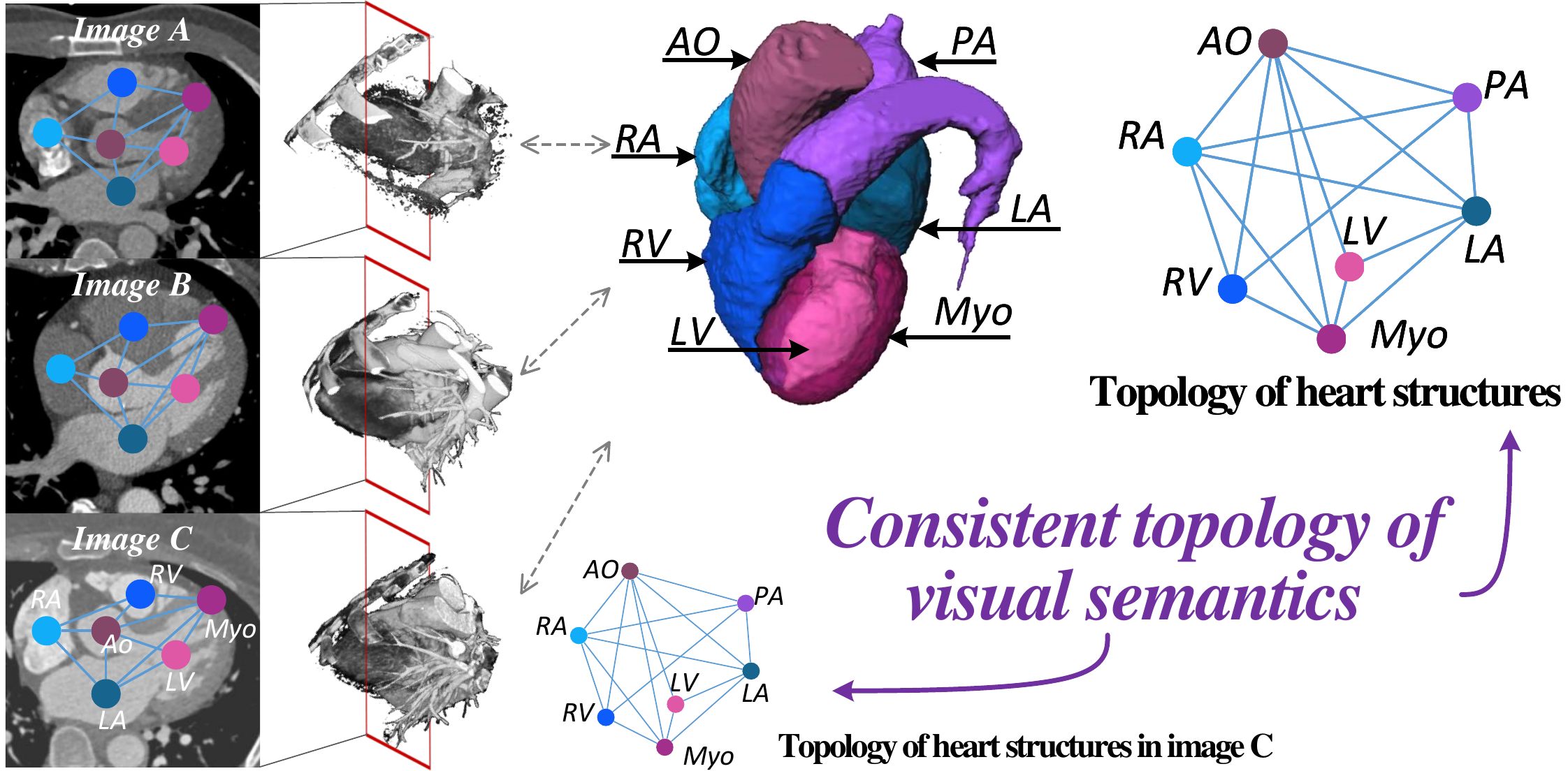}
\caption{The topological invariance of the visual semantics between the 3D medical images provides a motivation to discover their inter-image correspondence.}
\label{fig:motivation}
\end{figure}

Topological invariance \cite{heimann2009statistical,miller2001group} of the visual semantics in 3D medical images provides a motivation to construct a reliable measurement for inter-image similarity (Fig.\ref{fig:motivation}). Due to the consistency of human anatomies \cite{netter2014atlas}, 3D medical images have consistent context topology between the visual semantics in image space (e.g., the four chambers of human hearts have a fixed space relationship), and the same semantic regions have similar shapes in different images (e.g., the vessels (AO) have a stable tubular structure), constructing an invariant topology for the visual semantics. Therefore, according to the semantic prior of topological invariance, the semantic regions are able to be transformed to align in the image space via a topology-invariant mapping \cite{845378}, thus discovering their reliable inter-image correspondence even with large variations in appearance. An intuitive strategy is to use the registration or geometric matching (GM) methods \cite{he2021few,rocco2017convolutional,han2017scnet,haskins2020deep,he2022learning} to discover correspondence indexes between images, and use these indexes to constrain the consistent representation for corresponding regions. However, the errors in these indexes will bring mis-correspondence. 

In this paper, we propose a novel SSP paradigm, Geometric Visual Similarity Learning (GVSL), to learn the inter-image similarity in 3D medical images. It embeds the prior of topological invariance into the measurement of the similarities, and train network to estimate semantics' correspondence from the represented features in GM. Due to this effective semantic prior, the measurement will consider the semantic-related topology similarity avoiding the large interference of semantic-independent variation. Therefore, when learning to enlarge this similarity between two images for more accurate estimation of correspondence, the gradient in backpropagation will constrain the network to cluster the corresponding features in embedding space for more consistent representation. To drive the GM learning, we further propose a Z-Matching head to explore the global and local collaborative representation learning of inter-image similarity in our GVSL paradigm. It constructs a collaborative learning head with affine (global matching) and deformable (local matching) transformations \cite{haskins2020deep}, thus embedding the pre-trained model with a powerful transferring ability for potential downstream tasks.

Our contributions are summarized as follows: 1) Our work advances the learning of inter-image similarity in 3D medical image SSP, and pre-trains the network to learn a consistent representation for same visual semantics between images without annotation, pushing the representability of pre-trained models. 2) We propose the Geometric Visual Similarity Learning (GVSL) that embeds the prior of topological invariance into the metric for a reliable measurement of inter-image similarity, learning a consistent representation for same semantic regions between images. 3) We present a novel GM head, Z-Matching head, for simultaneously powerful global and local representation. It collaboratively learns the affine and deformable matching, realizing an effective optimization for the representation of different semantic granularity in our GVSL, and finally achieving a powerful transferring ability.  

\section{Related work}
\textbf{Learning similarity in self-supervised pre-training:} Learning similarity \cite{milbich2021visual,yan2022sam,you2022simcvd} targets on learning consistent representation for similar visual objects and distinguished representation for dissimilarity objects, which is a fundamental task in visual SSP \cite{jing2020self}. As illustrated in Sec.1, it has three main paradigms. Contrastive learning \cite{Chen2021CVPR,chen2020simple,He2020CVPR,NEURIPS20218757150d,you2022mine,you2023rethinking} which constrains the representation of same image to be consistent and different images to be separated. However, they are unable to learn the inter-images similarity, and the learning of separation will extremely interfere the representation of the 3D medical images. Clustering-based methods \cite{caron2018deep,li2021dense} are able to learn inter-image similarity, but their large-scale mis-measurement for similarities interferes the learning for consistent representation. Generation-based methods \cite{liu2021self,komodakis2018unsupervised,vincent2010stacked,zhou2019models,haghighi2020learning,haghighi2021transferable} generate pretext labels manually and constrain networks to predict these labels. However, it implicitly embeds bias from manual design into the learning which will make network ignore some potential aspects and limit in the representation in some specific scenario.

\textbf{Geometric matching \& Registration:} Geometric matching (GM, or named registration) \cite{he2021few,he2020deep,shi2022xmorpher,rocco2017convolutional,han2017scnet,haskins2020deep} aligns images' semantic regions to a same spatial coordinate system, thus providing correspondence indexes between two images. It has two level transformations: 1) Affine matching \cite{weisstein2004affine,haskins2020deep} aligns images in global. It calculates a transformation matrix that consists of the rotation, scaling, translation, and shearing operations between images and transforms the images to align in a global view. 2) Deformable matching \cite{he2021few,haskins2020deep,shi2022xmorpher,he2022learning} aligns images in local. It calculates a voxel-wise displacement vector field (DVF) which indicates the correspondence of the voxels between two images, and aligns the images via a spatial transformation operation \cite{jaderberg2015spatial}. Recently, due to the development of deep learning (DL), the DL-based GM \cite{rocco2017convolutional,he2021few,haskins2020deep} learns the representation driven by learning correspondence prediction, which provides us a potential solution.

\begin{figure*}[htb]
\centering
\includegraphics[width=\linewidth]{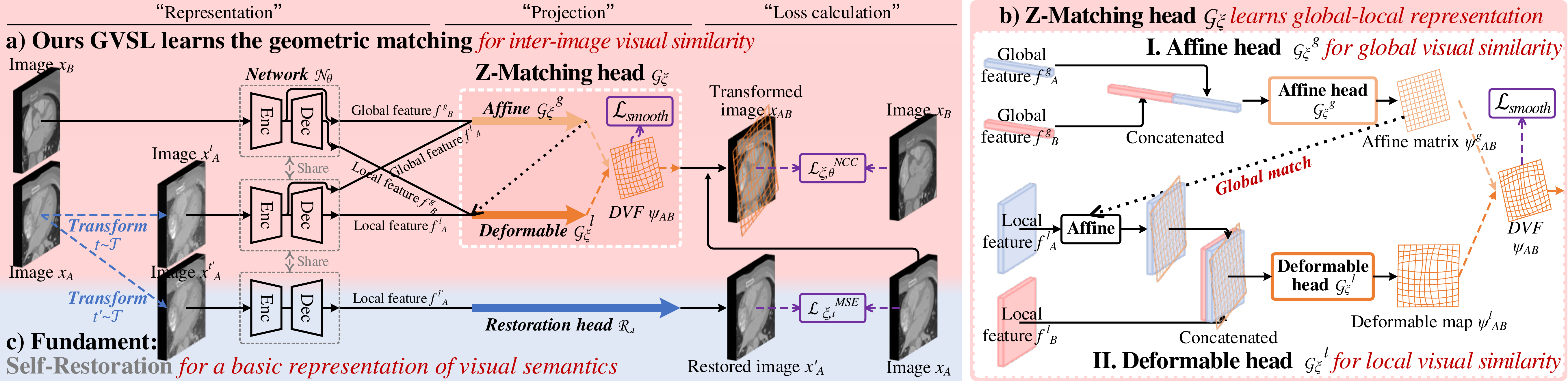}
\caption{The framework of our GVSL: \textbf{a)} Our GVSL learns the GM from the representation of the semantics in images, thus driving the learning of inter-image similarity via the gradient in backpropagation. \textbf{b)} Our Z-Matching head learns affine and deformable matchings simultaneously for powerful global and local representations. \textbf{c)} For efficient learning, it also takes a fundamental pretext task, the self-restoration, for a basic representation of semantics, thus giving a warm-up for GM learning.}
\label{fig:framework}
\end{figure*}
\section{Methodology}
Our framework (Fig.\ref{fig:framework}) learns from scratch on unlabeled 3D medical images, yielding a common visual representation with inter-image similarity. 
\subsection{GVSL for inter-image similarity}
\label{sec:sgm}
The proposed GVSL (Fig.\ref{fig:framework} a)) models the learning of inter-image similarity as the estimation of inter-image correspondence from represented features which embeds the prior of topological invariance into the measurement, thus utilizing the gradient in backpropagation to train the network to represent consistent features on same semantics.

\textbf{3.1.1 Description of GVSL}
GVSL's goal is to learn a representation that has a powerful clustering effect of the same semantic features even in different images. As shown in Fig.\ref{fig:framework}, it uses two shared-weight neural networks $\mathcal{N}_{\theta}$ with weights $\theta$ to represent the features $f_{A},f_{B}$ from two images $x_{A},x_{B}$. These features are put into a GM head $\mathcal{G}_{\xi}$ (our Z-matching head in our framework, Sec.3.3\ref{sec:gm}), to learn the correspondence of the semantic regions between images, thus driving the consistent representation in $\mathcal{N}_{\theta}$ for these same semantics.

Given a set of 3D medical images $\mathcal{D}$, two images $x_{A},x_{B}\sim\mathcal{D}$ are sampled uniformly from $\mathcal{D}$ ($A$ and $B$ refer to different images), and one transformation operation $t\sim\mathcal{T}$ is sampled from a transformation set $\mathcal{T}$. GVSL produce a transformed view $x^{t}_{A}\triangleq t(x_{A})$ from $x_{A}$ by applying the transformation $t$ to improve the diversity of the images. The network outputs global and local representations $\{f^{g}_{A},f^{l}_{A}\}\triangleq \mathcal{N}_{\theta}(x^{t}_{A}), \{f^{g}_{B},f^{l}_{B}\}\triangleq \mathcal{N}_{\theta}(x_{B})$ of $x^{t}_{A}$ and $x_{B}$. The head for GM $\mathcal{G}_{\xi}$ further outputs a displacement vector field (DVF) $\psi_{AB}\triangleq \mathcal{G}_{\xi}(f^{l}_{A},f^{l}_{B},f^{g}_{A},f^{g}_{B})$ which indicates the correspondence of the voxels between two images. The spatial coordinate system of the image $x_{A}$ is transformed to the image $x_{B}$ for a geometric matched image $x_{AB}\triangleq \psi_{AB}(x_{A})$ via spatial transformation \cite{jaderberg2015spatial}. We utilize the in-painting, local-shuffling, and non-linear as the transformation set $\mathcal{T}$.

We calculate a normalized cross-correlation (NCC) to evaluate the alignment degree between two images which indirectly evaluates the accuracy of the predicted correspondence. We train the framework to minimize this loss $\mathcal{L}^{NCC}$, thus driving the learning of correspondence,
\begin{align}\label{equ:ncc}
&\mathcal{L}_{\theta, \xi}^{NCC}\triangleq\\
&\sum_{p\in\Omega}\frac{(\sum_{p_{i}}(x_{B}(p_{i})-\hat{x_{B}}(p))(x_{AB}(p_{i})-\hat{x_{AB}}(p)))^{2}}{(\sum_{p_{i}}(x_{B}(p_{i})-\hat{x_{B}}(p))^{2})(\sum_{p_{i}}(x_{AB}(p_{i})-\hat{x_{AB}}(p))^{2})},\notag
\end{align}
where the $p$ is the position of the voxels in the image space $\Omega$, and the $i$ is the index of $p$. The $\hat{x_{B}}(p)$ is the local mean intensity images: $\hat{x_{B}}(p)=\frac{1}{n^{3}}\sum_{p_{i}}x_{B}(p_{i})$ (the same as $\hat{x_{AB}}(p)$). To keep the topology invariant in GM, we further calculate a smooth loss $\mathcal{L}^{smooth}$ on the DVF, constraining the network to perceive the correspondence of semantic regions under the condition of their topology,
\begin{equation}\label{equ:smooth}
\mathcal{L}_{\theta, \xi}^{smooth}\triangleq\sum_{p\in\Omega}\|\bigtriangledown \psi_{AB}(p)\|^{2}.
\end{equation}

At each training step, we perform a stochastic optimization step to minimize $\mathcal{L}^{GVSL}_{\theta,\xi}=\mathcal{L}_{\theta,\xi}^{NCC}+\mathcal{L}_{\theta,\xi}^{smooth}$ with the weights of $\theta$ and $\xi$. Therefore, the framework's dynamics are summarized as $\{\theta,\xi\}\leftarrow optimizer(\{\theta,\xi\}, \nabla_{\theta,\xi}\mathcal{L}^{GVSL}_{\theta,\xi}, \eta)$, where the $optimizer$ is an optimizer in training and $\eta$ is a learning rate.

\textbf{3.1.2 Intuitions on GVSL’s behavior}\label{sec:gm}
\begin{figure}
\centering
\includegraphics[width=\linewidth]{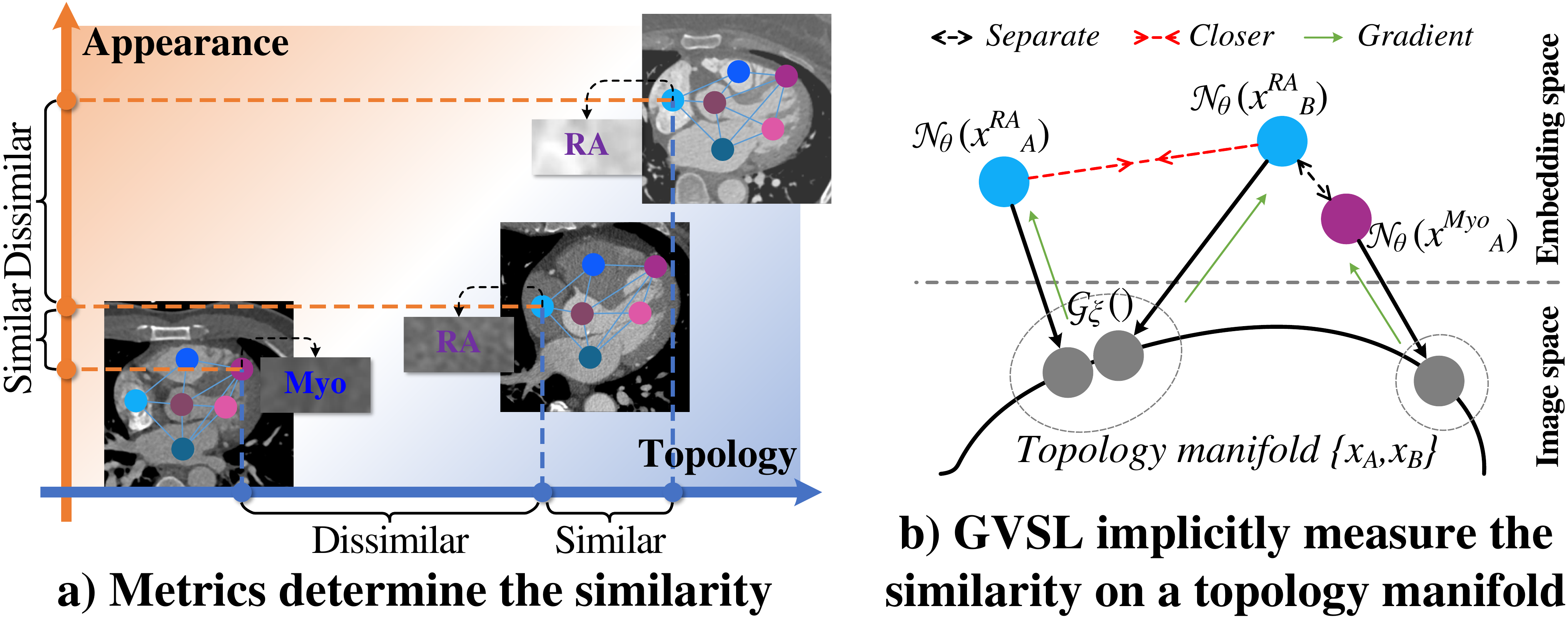}
\caption{Intuitions on GVSL’s behavior: The prior of topological invariance in GM embeds a topology manifold into the metric, thus bringing an efficient measurement for inter-image similarity and guiding the clustering effect of same semantic features.}
\label{fig:metric}
\end{figure}
The GM head together with the loss for similarity ($\mathcal{L}^{NCC}$) and the loss for topology-preservation ($\mathcal{L}^{smooth}$) in our GVSL is an implicit metric \cite{bellet2015metric} with the prior of topological invariance. As shown in Fig.\ref{fig:metric} a), metrics determine the similarity between images. The GM limits the measurement of inter-image visual similarity under the condition of the invariant topology in the image space, avoiding mis-measurement caused by the appearance (Fig.\ref{fig:limitation}).
\begin{equation}\label{equ:behavior}
\min_{\theta,\xi}\mathcal{L}(\mathcal{G}_{\xi}(\mathcal{N}_{\theta}(x_{A}),\mathcal{N}_{\theta}(x_{B}));\{x_{A},x_{B}\})
\end{equation}
It implicitly embeds a topology manifold inner the images $\{x_{A},x_{B}\}$ into the measurement process, and measure the similarity ($\mathcal{L}^{NCC}$) on this topology manifold (Fig.\ref{fig:metric} b), Equ.\ref{equ:behavior}). The $x_{A}^{RA}$, $x_{B}^{RA}$ and $x_{A}^{Myo}$ are the potential RA regions on images $x_{A},x_{B}$ and the potential Myo region on image $x_{A}$. In embedding space, due to the similarity in appearance, the distance between the features of RA in image B $\mathcal{N}_{\theta}(x^{RA}_{B})$ and the features of Myo in image A $\mathcal{N}_{\theta}(x^{Myo}_{A})$ is closer than that between the features of RA regions in image A $\mathcal{N}_{\theta}(x^{RA}_{A})$ and B $\mathcal{N}_{\theta}(x^{RA}_{B})$. This will bring mis-correspondence via some direct metrics, e.g., Euclidean distance. The GM $\mathcal{G}_{\xi}$ in our GVSL maps these represented features ($\mathcal{N}_{\theta}(x_{A}^{RA}), \mathcal{N}_{\theta}(x_{B}^{RA}), \mathcal{N}_{\theta}(x_{A}^{Myo})$) in the embedding space to the topology manifold inner the images $\{x_{A},x_{B}\}$ in image space. Therefore, due to the prior of the topological invariance, the distance between the RA regions will be closer than that between the $x_{B}^{RA}$ and $x_{A}^{Myo}$ and bring efficient learning of inter-image similarity via the gradient, thus learning efficient clustering effect.

Throughout the whole training process, the learning of representation for inter-image similarity in the network $\mathcal{N}_{\theta}$ and the correspondence in the GM head $\mathcal{G}_{\xi}$ is a two-player game \cite{saad2009coalitional}. The GM head $\mathcal{G}_{\xi}$ learns to estimate the correspondence of semantic regions from the represented features $f_{A},f_{B}$ and measure their voxel displacement in image space. The network $\mathcal{N}_{\theta}$ learns to provide features of visual semantic regions to the GM head $\mathcal{G}_{\xi}$ for their correspondence. To achieve more accurate correspondence, the GM head has to drive the pre-training network to output more consistent and representative features in turn for same semantic regions via the gradient in backpropagation. Therefore, under this interaction, the network $\mathcal{N}_{\theta}$ will provide more representative features for the GM head for better correspondence estimation, and the GM head $\mathcal{G}_{\xi}$ will have a more powerful ability to learn the inter-image similarity.

\subsection{Z-Matching for Global-Local Representations}\label{sec:zm}
The proposed Z-Matching $\mathcal{G}_{\xi}$ (Fig.\ref{fig:framework} b)) is a novel GM head that collaboratively learns affine and deformable matching for simultaneous global and local representation. It has two sub-head including the affine head $\mathcal{G}^{g}_{\xi}$ for global similarity and the deformable head $\mathcal{G}^{l}_{\xi}$ for local similarity.

\textbf{3.2.1 Affine head for global visual similarity}
It concatenates the global features ($f^{g}_{A},f^{g}_{B}$) of two images from the encoder part of the network $\mathcal{N}_{\theta}$, and puts these features into the affine head $\mathcal{G}^{g}_{\xi}$ to predict the affine values. 15 values (3 for rotation ($\theta_x,\theta_y,\theta_z$), 3 for translation ($t_x,t_y,t_z$), 3 for scaling ($s_x,s_y,s_z$), and 6 for shearing ($sh_{xy},sh_{xz},sh_{yx},sh_{yz},sh_{zx},sh_{zy}$)) are estimated to calculate the affine matrix $\psi^{g}_{AB}\triangleq \mathcal{G}^{g}_{\xi}(f^{g}_{A},f^{g}_{B})$ which indicates the global transformation target in the spatial coordinate system of $x_{A}$ to align the $x_{B}$ in global. Therefore, to percept the global correspondence, the optimizer will constrain the encoder to extract consistent and representative features for same global semantic regions.

\textbf{3.2.2 Deformable head for local visual similarity}
It takes the affine matrix $\psi^{g}_{AB}$ to transform the global spatial coordinate system of the local features (from the decoder of the network $\mathcal{N}_{\theta}$) $f^{l}_{A}$ to the $f^{l}_{B}$ for a global matching, thus the local features are globally aligned. It further concatenates the local features and puts them into the deformable head $\mathcal{G}^{l}_{\xi}$ to predict a deformable map $\psi^{l}_{AB}\triangleq\mathcal{G}^{l}_{\xi}(\psi^{g}_{AB}(f^{l}_{A}),f^{l}_{B})$ that will deform the voxels in the image $A$ to align their corresponding voxels in image $B$. Therefore, to achieve this voxel-wise alignment, the optimizer will constrain the whole network to extract consistent and representative features for same local visual semantic regions. Finally, the affine matrix $\psi^{g}_{AB}$ and the deformable map $\psi^{l}_{AB}$ are fused ($\odot$) for the DVF $\psi_{AB}\triangleq\psi^{l}_{AB}\odot\psi^{g}_{AB}$. (Details of $\odot$ are in \emph{Supplementary Material}.)

Therefore, the correspondence of the visual semantic regions between two images is predicted in $\psi_{AB}$, and the learning of correspondence will constrain the network $\mathcal{N}_{\theta}$ to extract more consistent and representative features for same visual semantics, thus driving the head $\mathcal{G}_{\xi}$ to have more powerful ability to discover their correspondence.

\subsection{Fundamental pretext task for warm-up}
The initial basic representation for visual semantics is important in our GVSL, so that we utilize self-restoration \cite{zhou2019models} as the fundamental pretext task (fundament) in our framework. During the learning of GM, the learning of correspondence $\mathcal{G}_{\xi}$ relies on the represented features of two images from the network $\mathcal{N}_{\theta}$. The initial network $\mathcal{N}_{\theta}$ with weak representability will limit the discovery of the correspondence between potential visual semantics, making it challenging to find a reliable optimization target to align same semantic regions, hindering the inter-image similarity learning. Therefore we construct a self-restoration \cite{zhou2019models} task in the framework for a warm-up of the GVSL.

As shown in Fig.\ref{fig:framework} c), it randomly transforms the appearance of an image ($x_{A}$) via a sampled transformation operation ($t'\sim\mathcal{T}$) for a transformed image ($x^{t'}_{A}\triangleq t'(x_{A})$) and put it into the network $\mathcal{N}_{\theta}$ to represent its local feature $f^{l'}_{A}$ from the decoder. (To save computing resources, we share this operation with GVSL in our experiment, i.e., $\mathcal{N}_{\theta}(x^{t'}_{A})=\mathcal{N}_{\theta}(x^{t}_{A})$.) Then, the feature $f^{l'}_{A}$ is put into a restoration head $\mathcal{R}_{\iota}$ for a restored image ($x'_{A}$), and calculates mean sequence error losses $\mathcal{L}^{MSE}_{\theta,\iota}=\|\mathcal{R}_{\iota}(\mathcal{N}_{\theta}(t'(x_{A})))-x_{A}\|^{2}$ \cite{zhou2019models} with the original image $x_{A}$ to learn the restoration of the visual semantics from a transformed context. Therefore, the network will learn a basic representation of semantics for warm-up in dynamics, i.e., $\{\theta,\iota\}\leftarrow optimizer(\{\theta,\iota\}, \nabla_{\theta,\iota}\mathcal{L}^{MSE}_{\theta,\iota}, \eta)$, avoiding the weak optimization in GVSL.
\begin{table*}[htb]
\centering
\caption{The linear (a) and fine-tuning (b) evaluations demonstrate our powerful representation and great transferring ability. The cells with a pink background are the top value in the columns. The red or blue values are the improvement or reduction compared with the ``Scratch".}
\resizebox{\textwidth}{!}{
\begin{tabular}{|l|ccc|c|ccc|c|}
\hline
\multirow{2}{*}{Pre-training}&\multicolumn{4}{c|}{\textbf{a) Linear: powerful representation}}&\multicolumn{4}{c|}{\textbf{b) Fine-tuning: great transferring}}\\
\cline{2-9}
                         &SHC$_{DSC\%}$&SAC$_{DSC\%}$&CCC$_{AUC\%}$&SBM$_{DSC\%}$&SHC$_{DSC\%}$&SAC$_{DSC\%}$&CCC$_{AUC\%}$&SBM$_{DSC\%}$\\
                         \cline{2-9}
                         &\multicolumn{3}{c|}{\emph{Inner scene}}&\emph{Inter scene}&\multicolumn{3}{c|}{\emph{Inner scene}}&\emph{Inter scene}\\
\hline
Scratch                              &21.9  &10.0&52.7&56.4&87.8   &80.4&74.4 &89.7\\
\hline
Denosing \cite{vincent2010stacked}    &31.4$_{\color{red}(+9.5)}$  &9.3$_{\color{blue}(-0.7)}$&57.9$_{\color{red}(+5.2)}$&28.3$_{\color{blue}(-28.1)}$&90.3$_{\color{red}(+2.5)}$ &80.5 $_{\color{red}(+0.1)}$  &75.6$_{\color{red}(+1.2)}$&89.7\\
In-painting \cite{pathak2016context}  &32.3$_{\color{red}(+10.4)}$ &5.9$_{\color{blue}(-4.1)}$&57.1$_{\color{red}(+4.4)}$&25.0$_{\color{blue}(-31.4)}$&90.4$_{\color{red}(+2.6)}$ &80.3 $_{\color{blue}(-0.1)}$ &79.9$_{\color{red}(+5.5)}$&89.9$_{\color{red}(+0.2)}$\\
Models Genesis \cite{zhou2019models}  &47.4$_{\color{red}(+25.5)}$ &22.5$_{\color{red}(+12.5)}$&60.4$_{\color{red}(+7.7)}$&44.9$_{\color{blue}(-11.5)}$&90.3$_{\color{red}(+2.5)}$ &79.9 $_{\color{blue}(-0.5)}$ &80.7$_{\color{red}(+6.3)}$&89.4$_{\color{blue}(-0.3)}$\\
Rotation \cite{komodakis2018unsupervised}&56.1$_{\color{red}(+34.2)}$&21.9$_{\color{red}(+11.9)}$&\cellcolor[HTML]{FFCCC9}\textbf{62.1$_{\color{red}(+9.4)}$}&54.1$_{\color{blue}(-2.3)}$&90.6$_{\color{red}(+2.8)}$&81.1$_{\color{red}(+0.7)}$&77.1$_{\color{red}(+2.7)}$&89.6$_{\color{blue}(-0.1)}$\\
DeepCluster \cite{caron2018deep}      &55.9$_{\color{red}(+34.0)}$&4.4$_{\color{blue}(-5.6)}$&57.9$_{\color{red}(+5.2)}$&67.5$_{\color{red}(+11.1)}$&85.4$_{\color{blue}(-2.4)}$&80.5$_{\color{red}(+0.1)}$&59.9$_{\color{blue}(-14.5)}$&89.1$_{\color{blue}(-0.6)}$\\
SimSiam \cite{Chen2021CVPR}      &56.5$_{\color{red}(+34.6)}$ &9.7$_{\color{blue}(-0.3)}$&61.0$_{\color{red}(+8.3)}$&66.2$_{\color{red}(+9.8)}$&87.5$_{\color{blue}(-0.3)}$&80.1 $_{\color{blue}(-0.3)}$ &73.6$_{\color{blue}(-0.8)}$&89.8$_{\color{red}(+0.1)}$\\
BYOL \cite{grill2020bootstrap}   &46.9$_{\color{red}(+25.0)}$ &8.6$_{\color{blue}(-1.4)}$&53.7$_{\color{red}(+1.0)}$&52.7$_{\color{blue}(-3.7)}$&88.6$_{\color{red}(+0.8)} $&80.7 $_{\color{red}(+0.3)}$  &76.5$_{\color{red}(+2.1)}$&89.5$_{\color{blue}(-0.2)}$\\
SimCLR \cite{chen2020simple}      &48.7$_{\color{red}(+26.8)}$ &15.5$_{\color{red}(+5.5)}$&61.3$_{\color{red}(+8.6)}$&58.7$_{\color{red}(+2.3)}$&86.9 $_{\color{blue}(-0.9)}$ &79.9$_{\color{blue}(-0.5)}$ &74.3$_{\color{blue}(-0.1)}$&89.3$_{\color{blue}(-0.4)}$\\
\hline
w/o Z-Matching     &49.1$_{\color{red}(+27.2)}$ &21.1$_{\color{red}(+11.1)}$&55.8$_{\color{red}(+3.4)}$&45.1$_{\color{blue}(-11.3)}$&88.3$_{\color{red}(+0.5)}$ &81.2$_{\color{red}(+0.8)}$&81.3$_{\color{red}(+6.9)}$ &89.7 \\
w/o Fundament               &45.3$_{\color{red}(+23.4)}$ &0.0$_{\color{blue}(-10.0)}$&58.8$_{\color{red}(+6.4)}$&48.5$_{\color{blue}(-7.9)}$&87.0$_{\color{blue}(-0.8)}$&79.5 $_{\color{blue}(-0.9)}$ &76.6$_{\color{red}(+2.2)}$&89.0$_{\color{blue}(-0.7)}$\\
w/o Affine head        &57.7$_{\color{red}(+35.8)}$&17.9$_{\color{red}(+7.9)}$&57.6$_{\color{red}(+4.9)}$&53.4$_{\color{blue}(-3.0)}$&89.4$_{\color{red}(+1.6)}$&\cellcolor[HTML]{FFCCC9}\textbf{82.3$_{\color{red}(+1.9)}$}&79.8$_{\color{red}(+5.4)}$&89.8$_{\color{red}(+0.1)}$\\
\textbf{Our GVSL (Whole)}                 &\cellcolor[HTML]{FFCCC9}\textbf{68.4$_{\color{red}(+46.5)}$}&\cellcolor[HTML]{FFCCC9}\textbf{28.7$_{\color{red}(+18.7)}$}&60.8$_{\color{red}(+8.1)}$&\cellcolor[HTML]{FFCCC9}\textbf{79.9$_{\color{red}(+23.5)}$} &\cellcolor[HTML]{FFCCC9}\textbf{91.2$_{\color{red}(+3.4)}$}&81.3$_{\color{red}(+0.9)}$&\cellcolor[HTML]{FFCCC9}\textbf{82.2$_{\color{red}(+7.8)}$}&\cellcolor[HTML]{FFCCC9}\textbf{90.0$_{\color{red}(+0.3)}$}\\
\hline
\end{tabular}
}
\label{tab1}
\end{table*}
\section{Experiments and Results}
\subsection{Experiment protocol}
\textbf{1) Materials:} Five datasets are used in our experiments. \textbf{a)} Pre-training dataset: Cardiac CT images from 302 patients are used as the self-supervised pre-training dataset without annotations. These images were acquired on a SOMATOM Definition Flash and the contrast media was injected during the CT image acquisition. The x/y-resolution of these CT images is between 0.25 to 0.57 mm/voxel and the slice thickness is between 0.75 to 3 mm/voxel. The x/y-size of the images is 512 voxels and the z-size is between 128  to 994 voxels. \textbf{b)} Downstream datasets: Four public available datasets (MM-WHS-CT \cite{zhuang2019evaluation}, ASOCA \cite{gharleghi2022automated}, CANDI \cite{kennedy2012candishare}, STOIC \cite{revel2021study}) are used to demonstrate the superiorities of our framework. According to their data kinds, we use them for inner-scene and inter-scene evaluations. For \textbf{inner-scene} evaluation, it utilizes the \emph{S}egmentation of seven \emph{H}eart structures on cardiac \emph{C}T images (SHC) \cite{zhuang2019evaluation}, \emph{S}egmentation of coronary \emph{A}rtery on cardiac \emph{C}T images (SAC) \cite{gharleghi2022automated}, and diagnosis (\emph{C}lassification) of \emph{C}OVID-19 on chest \emph{C}T images (CCC) \cite{revel2021study} to evaluate the adaptability for \emph{same} scenes (Cardiac or chest CT) as the source dataset. For \textbf{inter-scene} evaluation, it utilizes the \emph{S}egmentation of 28 \emph{B}rain tissues on \emph{M}R images (SBM) \cite{kennedy2012candishare} to evaluate the adaptability for \emph{different} scenes (Brain MR) as the source dataset. More details are in our \emph{Supplementary Material}.

\textbf{2) Comparisons:} We take eight works to benchmark our framework, including the generation-based methods \cite{vincent2010stacked,pathak2016context,zhou2019models,komodakis2018unsupervised} and contrast-based methods \cite{caron2018deep,chen2020simple,grill2020bootstrap,Chen2021CVPR}. Therefore, the superiority of our GVSL will be demonstrated. We take 3D U-Net \cite{cciccek20163d} as the backbone network for all methods (the global prediction methods use the encoder part of the backbone) for a fair comparison and use both fine-tuning and linear evaluations for a comprehensive demonstration.

\textbf{3) Evaluation metrics:} We use the mean Dice coefficient (DSC) to evaluate the segmentation tasks, and the Area Under the Curve (AUC) to evaluate the classification task following \cite{taha2015metrics}.

\textbf{4) Implementation:} All tasks are implemented by PyTorch \cite{paszke2019pytorch} on NVIDIA GeForce RTX 3090 GPUs with 24 GB memory, optimized by Adam \cite{kingma2014adam} whose learning rate is $10^{-4}$. The pretext task is trained with $2\times10^{5}$ iterations. The downstream tasks are trained with $4\times10^{4}$ iterations and validated every 200 iterations to save the best models on their validation sets. For a fair comparison, all methods in our experiment take the same basic training setting.
\subsection{Comparison study shows our superiority}
\begin{figure*}
\centering
\includegraphics[width=0.9\linewidth]{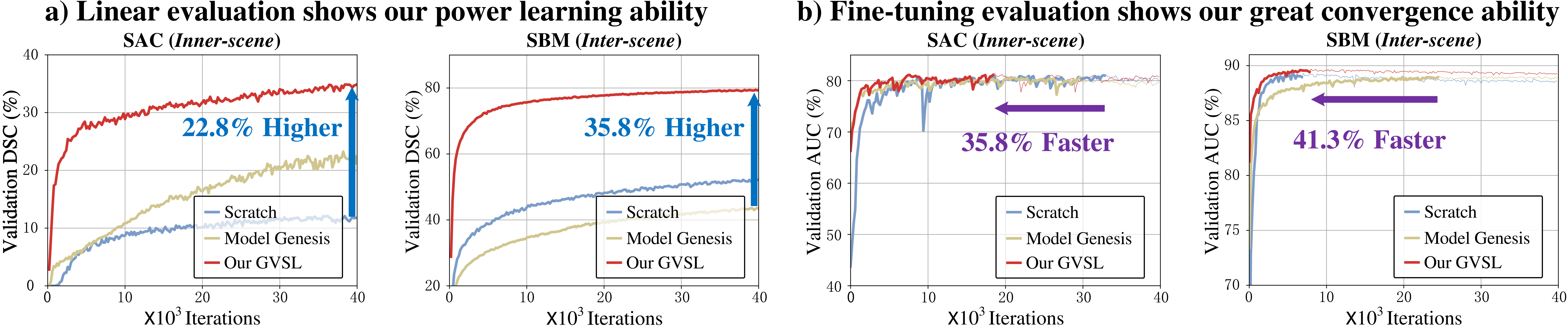}
\caption{Our GVSL has powerful representability in linear evaluation and faster convergence ability in fine-tuning evaluation. In b), the thick part of the lines mean the useful training process, and at the end of the thick parts are the saved best models on validation set.}
\label{fig:ana}
\end{figure*}
As demonstrated in Tab.\ref{tab1}, our linear (a) and fine-tuning (b) evaluations demonstrate our power representation ability and great transferring ability.

\textbf{4.2.1 Powerful \emph{inner-scene} transferring for both large and small structures} Our powerful inner-scene transferring ability shows the great application potential of our GVSL in big-data but low-label scenarios of medical images. It achieves the highest performance both in large and small structures in the same scene of the pre-training dataset. \textbf{1)} Large structures: In the SHC task which segments large cardiac structures, our GVSL achieves the highest DSC (68.4\%) in linear evaluation which is 11.9\% higher than the second-highest method \cite{Chen2021CVPR}. This is because our learning of inter-image similarity promotes the representation of consistent features. Especially for the large anatomies which have clear visual semantics, our GM brings a much more efficient representation. \textbf{2)} Small structures: In the SAC task which segments the small coronary arteries, numerous pre-training methods \cite{pathak2016context,Chen2021CVPR,grill2020bootstrap,chen2020simple} mislead the model to ignore such small features, resulting in even lower performance than the ``Scratch", especial for those methods \cite{Chen2021CVPR,grill2020bootstrap,chen2020simple} designed for global prediction. Our learning of the deformable transformation and the self-restoration teaches the pre-trained network consistent and effective representation for small visual semantics, achieving the highest DSC in linear (28.7\%) evaluation. It is also interesting that when removing the affine head in our Z-matching head, our GVSL achieves the best fine-tuning performance (82.3\%). This further demonstrates the importance of learning dense representation for small objects.

\textbf{4.2.2 Effective \emph{inter-scene} transferring} Our effective inter-scene transferring ability demonstrates our superiority as the initiation for deep networks. The SBM task uses brain MR images which have different modality and context (body range) with the pre-training dataset, making a challenging inter-scene transferring. In linear evaluation, a lot of compared methods \cite{zhou2019models,grill2020bootstrap,komodakis2018unsupervised,grill2020bootstrap} are unable to bring promotion in this task and achieves even worse performance than ``Scratch" due to the large difference between the source (brain MR) and target (cardiac CT) tasks. Our framework which is pre-trained on cardiac CT images is able to efficiently adapt to the segmentation task on brain MR images. Therefore, it achieves the highest 79.9\% DSC (23.5\% improvement). This is because the inter-image similarity brings the pre-trained network a better clustering effect for same semantic features even in different images, making the representation easier to be transferred to the target scene integrally. It is worth noting that in the fine-tuning evaluation, all methods only have similar even worse \cite{chen2020simple,grill2020bootstrap,caron2018deep,komodakis2018unsupervised,zhou2019models} performance as the ``Scratch". This is because the large distribution gap between brain MR and cardiac CT images (different modalities and body ranges) makes the pre-trained representability unable to achieve valid transferring. Our GVSL still achieves the highest 0.3\% improvement.

\textbf{4.2.3 Superiority in \emph{global} and \emph{dense} prediction tasks} Our superiority in both global and dense prediction tasks shows its great adaptability to potential downstream tasks. \textbf{1)} Dense prediction tasks (SHC, SAC, SBM): Our GVSL has the highest DSC for all three tasks owing to our deformable head and self-restoration head which learns representative and consistent features extraction ability for details. The SimSiam, BYOL, SimCLR, and DeepCluster only learn global representation in their pretext tasks, having very poor performance in the A.CT.S which focuses on detail features. \textbf{2)} Global prediction tasks (CCC): The medical images are similar globally and their lesions are on the local regions. Our GVSL utilizes our Z-Matching head to simultaneously learn the global and local visual similarity for global-local representation, achieving the highest AUC (82.2\%) in the fine-tuning evaluation and illustrating our superiority in the transferring of global prediction tasks. Although our GVSL has 60.8\% AUC in linear evaluation which is 1.3\% lower than the highest method (Rotation), it is still higher than the Denoising, In-painting, and Model Genesis which pre-train network via dense prediction tasks.
\begin{figure*}
\centering
\includegraphics[width=0.8\textwidth]{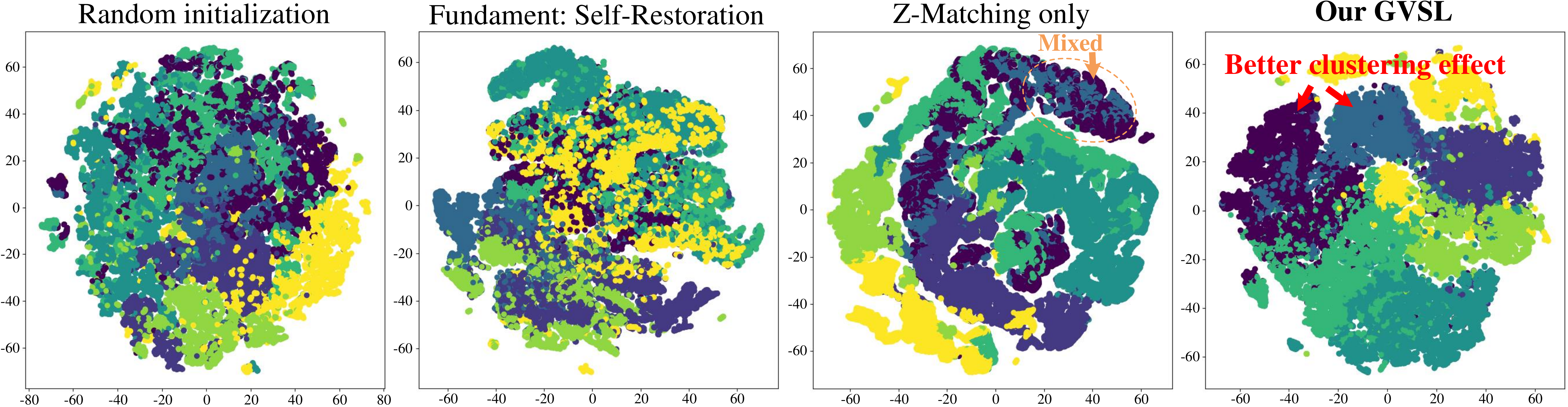}
\caption{Our GVSL's promotion for clustering effect. We draw the local features $f^{l}$ of seven semantic regions (AO, RA, RV, LV, PA, LA, and Myo) from the pre-trained network in the SHC task and compress them to 2 values by t-SNE \cite{van2008visualizing}. Our GVSL shows a great clustering effect for these semantics, while the features from other methods are mixed.}
\label{fig:clu}
\end{figure*}
\subsection{Ablation study and model analysis}
\textbf{4.3.1 Ablation study} We compare our GVSL with the only fundamental pretext task (self-restoration), the only Z-Matching for GM learning, and the fundament + only deformable matching, three observations can be found in Tab.\ref{tab1}. \textbf{1)} When only learning the GM (Z-Matching), its initial weak representability makes the pre-trained model have inefficient optimization and brings poor representation. Especially in the linear evaluation of the SAC task, it is unable to segment the extremely small structures due to the single GM's poor representation. \textbf{2)} When adding the fundamental task, our GVSL has better performance than the single two sub-pretext tasks on all four downstream tasks, showing the importance of the basic representation from self-restoration and the large contribution of our inter-image similarity from our GM. \textbf{3)} When removing the Affine head in the Z-Matching head, it reduces 3.2\% and 2.4\% AUC in the linear and fine-tuning evaluations of CCC task due to the lack of global representation learning. However, it achieves the highest DSC in the fine-tuning of the SAC task, because the targeted learning of deformable matching will promote the representation of thin structures in local features.

\textbf{4.3.2 Our promotion for the learning efficiency} As shown in Fig.\ref{fig:ana}, we analyze the learning of the models which are initialized from scratch, by our GVSL, and by the Model Genesis in the SAC and SBM tasks, demonstrating our powerful representability and much faster convergence ability. In the linear evaluation, our GVSL improves more than 20\% DSC compared with the 'scratch' or Model Genesis, owing to our effective learning for details in local-wise visual similarity. In the fine-tuning evaluation, our GVSL also greatly improves the convergence speed which achieves more than 30\% improvement, illustrating its great convergence ability and great potential for saving computing resources. Although in the fine-tuning of the SBM task, the "scratch" has faster convergence, it quickly falls into over-fitting, and its performance is extremely limited.
\begin{figure}
\centering
\includegraphics[width=0.8\linewidth]{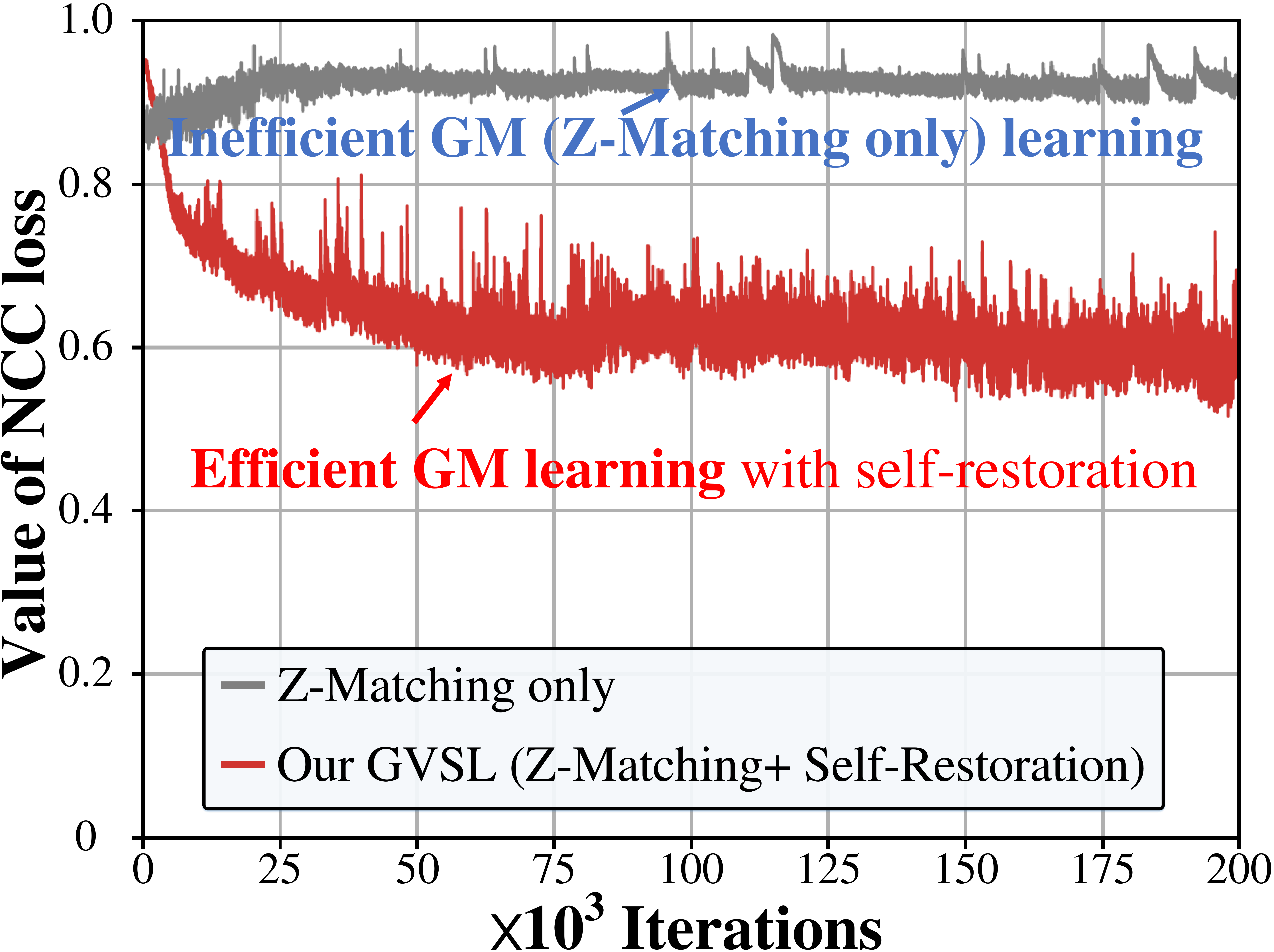}
\caption{The necessity of the fundament in our GM learning. a) When only training the GM task, the NCC loss $\mathcal{L}^{NCC}$ does not converge and is unable to learn the correspondence of semantic regions. b) When adding the fundament (self-restoration), warm-up from the basic representation of semantic regions drives the GM for efficient optimization.}
\label{fig:train}
\end{figure}

\textbf{4.3.3 The fundament's necessity for our GM learning} The self-restoration learns a basic representation for visual semantic regions, thus driving the learning of inter-image similarity in our GM. As demonstrated in Fig.\ref{fig:train}, when only learning our GM task, the network's initial weak representation makes inefficient optimization of the GM task, so the NCC loss $\mathcal{L}^{NCC}$ does not converge and is unable to learn the correspondence of semantic regions. When adding the fundamental pretext task, driven by the basic representation of visual semantic regions from the self-restoration, the NCC loss is successfully converging to learn the correspondence of semantic regions for a better clustering effect of same visual semantics inter images.

\textbf{4.3.4 Our GVSL's promotion for clustering effect} As shown in Fig.\ref{fig:clu}, the local features $f^{l}$ from the pre-trained models in the SHC task demonstrate our GVSL's promotion for the clustering effect. The random initialization mixes the different semantics, so it is unable to extract representative features and distinguish the potential visual semantics. When only learning the self-restoration, it learns basic representation for visual semantics, but is still limited by the constraint for inter-image similarity. Therefore, the local features $f^{l}$ are still mixed. When only learning our Z-Matching head for the GM task, the features have a significantly better clustering effect, showing the importance of inter-image similarity. However, the features of some semantics are still mixed. When the above two sub-pretext tasks are used simultaneously in our GVSL, the basic representation from the fundament promotes the learning of GM, so the clustering effect is more obvious and the features of different visual semantics are representative and distinguishing. Although the yellow points show three parts in our GVSL, each part is clustered, also showing that the representation of internal semantics of this region is consistent.
\section{Conclusion and Discussion}
In this paper, we have advanced the inter-image similarity learning in 3D medical image SSP, and proposed the \emph{Geometric Visual Similarity Learning (GVSL)} for the representation of inter-image similarity, achieving powerful representability for the transfer learning in downstream application-specific tasks. While its unique properties of learning consistent representation for same semantics have bright powerful performance in inner-scene, inter-scene, and global-local transferring tasks for 3D images (CT, MR), an important future work is to expand the learning of inter-image similarity to some images without topological invariance, i.e., whole slide imaging \cite{hanna2020whole}. We believe that our GVSL in SSP will promote the research of efficient learning in medical image analysis, and our GVSL is able to serve as a primary source of transfer learning for downstream tasks.

\textbf{Discussion for impact} The proposed method demonstrates an effective and reasonable potential in medical imaging analysis, showing great potential impact. Especially for the widely used 3D medical images, their \emph{spatial completeness} of 3D structures \cite{sabharwal2015completeness} avoids the spatial projection (e.g., X-ray images) and spatial occlusion (e.g., natural images) of 2D images. The consistency in human bodies also brings the \emph{topological invariance} of the content in these 3D images. Therefore, the geometric relationship between these images is able to be effectively used to drive the measurement of visual similarity. The spatial completeness and topological invariance of 3D structure in images will further inspire researchers to do more research on SSP.

\textbf{Discussion for limitation} There are still some limitations in our GVSL. 1) The additional calculation in the Z-Matching head and the fundamental self-restoration learning makes larger GPU memory requirement and more computing costs. 2) The inter-scene transferring is still a large challenge for medical image pre-training models due to the large gap between the source and target scenes. Fortunately, these limitations are gradually being solved due to the development of GPU and enlarging of medical datasets (provides more possibilities for inner-scene transferring).
\paragraph*{Acknowledgments}
This research was supported by the Intergovernmental Cooperation Project of the National Key Research and Development Program of China(2022YFE0116700), CAAI-Huawei MindSpore Open Fund and Scientific Research Foundation of Graduate School of Southeast University(YBPY2139). We thank the Big Data Computing Center of Southeast University for providing the facility support. We also thank the Key Laboratory of Computer Network and Information Integration, Ministry of Education, and the Jiangsu Provincial Joint International Research Laboratory of Medical Information Processing, Nanjing, China.

{\small
\bibliographystyle{ieee_fullname}
\bibliography{mybib}
}
\section*{Appendix A. Rethink GVSL and representation learning}
Our GVSL is an unsupervised representation learning paradigm which constructs a geometric metric to learn the inter-image similarity, thus achieving a consistent representation for same semantic regions based on a reliable semantics' correspondence.
\begin{figure}[htb]
  \centering
  \includegraphics[width=\linewidth]{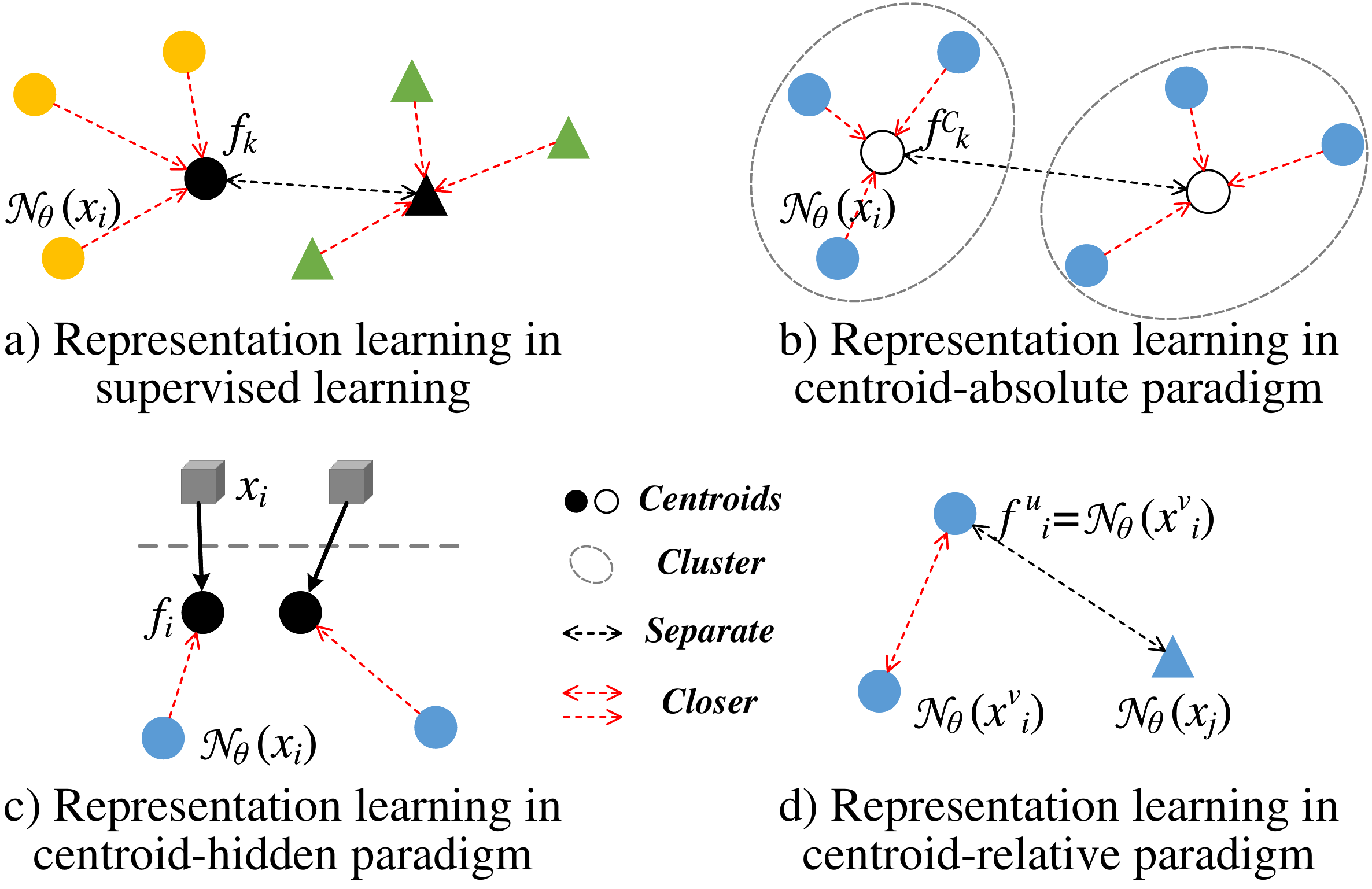}
  \caption{The view of the embedding space. a) The representation learning in supervised learning gather features to the centroids $\textbf{\text{f}}_{1:K}$ corresponding to their classes, and separate the centroids. b) Centroid-absolute paradigm clusters features for centroids $\textbf{\text{f}}^{\mathcal{C}}_{1:K}$, and learns to gather features to these centroids. c) Centroid-hidden paradigm generates the pretext labels via manual designed methods and learns follow the supervised learning. d) Centroid-relative paradigm train to gather the features of same image's different views, and separate the features of different images.}\label{supp:rep}
\end{figure}
\subsection*{A.1. Representation in supervised learning} Let's start by rethinking supervised learning from labeled dataset $\mathcal{D}=\{x_{i},y_{i}\}_{i=1}^{I},y_{i}\in\textbf{\text{y}}_{1:K}$, where $x_{i}$ and $y_{i}$ are the $i_{th}$ image and label, and $I$ is the number of data, $K$ is the number of classes. The whole framework can be divide into two parts, including the learning of representation $\mathcal{N}_{\theta}$ with parameters $\theta$ and the learning of specific task $\mathcal{G}_{\xi}$ with parameters $\xi$ \cite{goodfellow2016deep}. The representation part $\mathcal{N}_{\theta}$ maps images to an embedding space for features, and the specific task part $\mathcal{G}_{\xi}$ maps the features in the embedding space to the task space. The supervised learning train the network to learning the representation and specific task via minimizing the distance $d$ of framework's outputs and labels following
\begin{equation}\label{equ:1}
 \mathop{\min}_{\theta,\xi}d(\mathcal{G}_{\xi}(\mathcal{N}_{\theta}(x_{i})),y_{i})).
\end{equation}
We assume that there is a centroid $f_{k}\in\textbf{\text{f}}_{1:K}$ in the embedding space that makes $\mathcal{G}_{\xi}(f_{k})=y_{i}$, then the Equ.\ref{equ:1} is equivalent to
\begin{equation}\label{equ:2}
 \begin{aligned}
&\mathop{\min}_{\theta,\xi}d(\mathcal{N}_{\theta}(x_{i}),f_{k})\\
&\begin{array}{ll}
s.t. &\mathcal{G}_{\xi}(f_{k})=y_{i}.
\end{array}
\end{aligned}
\end{equation}
Obviously, in this process (Fig.\ref{supp:rep} a)), the representation part $\mathcal{N}_{\theta}$ is trained to gather features to the centroids $\textbf{\text{f}}_{1:K}$ corresponding to their classes via the specific task part $\mathcal{G}_{\xi}$. Therefore, the learning of $\mathcal{G}_{\xi}$ optimizes the centroids $\textbf{\text{f}}_{1:K}$ in the embedding space to distinguish different classes, and the learning of $\mathcal{N}_{\theta}$ optimizes the clustering effect of same class data.

\subsection*{A.2. Learning representation without annotation} When labels are unavailable $\mathcal{D}=\{x_{i}\}_{i=1}^{I}$, this means the centroids $\textbf{\text{f}}$ in the embedding space are unavailable to guide the clustering effect. Therefore, the self-supervised representation learning \cite{liu2021self} targets building the centroids $\textbf{\text{f}}$ via pretext tasks, thus guiding the network learning potential clustering effect. According to the difference of $\textbf{\text{f}}$, the existing methods can be divided into three paradigms:
\begin{itemize}
  \item Centroid-hidden paradigm (Fig.\ref{supp:rep} c)) \cite{liu2021self,komodakis2018unsupervised}: This paradigm still follows the Equ.\ref{equ:2}, and generates the pretext labels via designed transformation methods $\mathcal{T}$ (e.g., restoration \cite{zhou2019models}, rotation \cite{komodakis2018unsupervised}). Therefore, like the supervised learning, this paradigm impliedly creates centroids $\textbf{\text{f}}$ in the embedding space according to the pretext labels, learns the $\mathcal{N}_{\theta}$ to gather features to the centroids and learns the $\mathcal{G}_{\xi}$ to distinguish the centroids $\textbf{\text{f}}$ in embedding space.
      \begin{equation}\label{equ:3}
        \begin{aligned}
        &\mathop{\min}_{\theta,\xi}d(\mathcal{N}_{\theta}(x_{i}),f_{i})\\
        &\begin{array}{ll}
        s.t. &\mathcal{G}_{\xi}(f_{i})=\mathcal{T}(x_{i}).
        \end{array}
        \end{aligned}
        \end{equation}

      \emph{* Observation}: The centroids extremely depend on manual defined transformation methods $\mathcal{T}$, which will bring large bias in the representation. For example, the rotation method \cite{komodakis2018unsupervised} will make the $\mathcal{N}_{\theta}$ biased to the position features, and some images whose positions are semantics-independent information will be mis-represented.
  \item Centroid-absolute paradigm (Fig.\ref{supp:rep} b)) \cite{caron2018deep,li2021dense}: This paradigm utilizes the clustering methods $\mathcal{C}^{K}$ ($K$ is the number of clustered centroids) to discover the clustering patterns of features, thus building the centroids $\textbf{\text{f}}^{\mathcal{C}}_{1:K}$ and gathering the represented features to these centroids, like DeepCluster \cite{caron2018deep}
      \begin{equation}\label{equ:4}
         \begin{aligned}
        &\mathop{\min}_{\theta}d(\mathcal{N}_{\theta}(x_{i}),f^{\mathcal{C}}_{k})\\
        &\begin{array}{ll}
        s.t. &f^{\mathcal{C}}_{k}=\mathcal{C}^{K}(\mathcal{N}_{\theta}(x_{i});\mathcal{D}).
        \end{array}
        \end{aligned}
        \end{equation}

      \emph{* Observation}: The clustering method $\mathcal{C}^{K}$ is the bottleneck in this paradigm. The existing works \cite{caron2018deep,li2021dense} utilize K-means \cite{hamerly2003learning} as the clustering methods which is extremely interfered by images' semantic-independent variations. Therefore, the clustered centroids will bring imprecise information, finally learning mis-representation.

  \item Centroid-relative paradigm (Fig.\ref{supp:rep} d)) \cite{Chen2021CVPR,chen2020simple}: This paradigm has no explicit centroids $\textbf{\text{f}}$, but train $\mathcal{N}_{\theta}$ to contrast images. A popular method is contrastive learning \cite{Chen2021CVPR,chen2020simple,He2020CVPR}. This method constrains the representation of same image's different views ($x_{i}^{v},x_{i}^{u}$) to be consistent and different images ($x_{i},x_{j}$) to be separated, thus gaining clustering effect under the training of big data.
      \begin{equation}\label{equ:5}
             \begin{aligned}
            &\mathop{\min}_{\theta}d(\mathcal{N}_{\theta}(x_{i}^{v}),f_{i}^{u})-d(\mathcal{N}_{\theta}(x_{j}),f_{i}^{u})\\
            &\begin{array}{ll}
            s.t. &f_{i}^{u}\triangleq\mathcal{N}_{\theta}(x_{i}^{u})
            \end{array}
            \end{aligned}
      \end{equation}

        \emph{* Observation}: This paradigm have to learn inner-image similarity and inter-image dissimilarity. However, if the images share numerous same semantics, this paradigm will make the $\mathcal{N}_{\theta}$ learn the task-unconcerned features. Especially in our task, the 3D medical images share numerous same semantic regions due to the consistency of human anatomies, the direct learning of separation will mislead the consistent representation of these same semantic regions. Although some works have removed the learning of inter-image dissimilarity, the single learning of inner-image similarity will bring the risk of dimensional collapse \cite{jing2021understanding}.
  \end{itemize}

\textbf{Conclusion}: Observing these three paradigms, we can draw three conclusions:
\begin{itemize}
  \item \emph{A self-discovery method} to drive the learning of clustering effect is crucial to avoid the large bias caused by the manual designed transformation.
  \item \emph{Prior knowledge of semantics} is crucial to avoid the interference caused by images' semantic-independent variations during the self-discovery of clustering effect.
  \item \emph{Learning inter-image similarity} is crucial for 3D medical image self-supervised pre-training.
\end{itemize}

Therefore, our GVSL fuses the prior of topological invariance into the learning of inter-image similarity in a self-discovery process, achieving power self-supervised per-training.

\subsection*{A.3. Learning GVSL}
\begin{figure}
  \centering
  \includegraphics[width=0.8\linewidth]{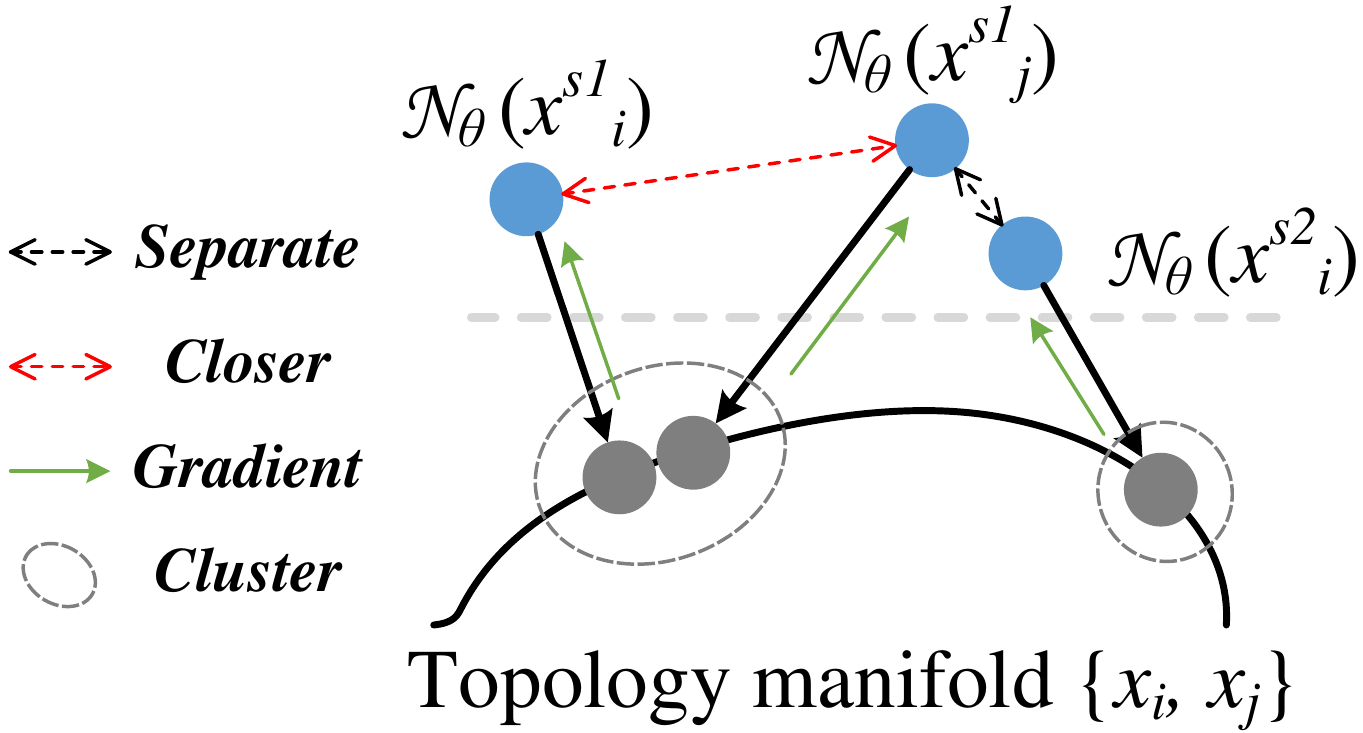}
  \caption{GVSL in the view of embedding space. It projects features onto a manifold of consistent topology, and gathers the semantic features ($\mathcal{N}_{\theta}(x^{s1}_{i}),\mathcal{N}_{\theta}(x^{s1}_{j})$, $s1$ means the semantic regions on images) which are closed on this manifold.}\label{supp:ours}
\end{figure}
Our GVSL embeds a geometric mapping into the measurement of different images, bringing three advancement compared with above three paradigms:
\begin{itemize}
  \item Compared with the centroid-hidden paradigm, it brings a self-discovery process which learns a geometric matching head $\mathcal{G}_{\xi}$ to discover the corresponding of visual objects between images to learn consistent representation of same semantics.
  \item Compared with the centroid-absolute paradigm, it embeds the prior of topological invariance into the discovery of correspondence, avoiding the interference caused by images' semantic-independent variations.
  \item Compared with the centroid-relative paradigm, it avoids the direct learning of inter-image dissimilarity in global, and utilizes the geometric matching to discover the correspondence of same semantic regions inner two images and learn consistent representation of them.
\end{itemize}

Compared with the Equ.\ref{equ:5}, GVSL (Equ.\ref{equ:6}) takes a $\mathcal{G}_{\xi}$ to discover the correspondence of same semantic regions between two images, avoiding the direct enlarging of feature distance for two images in Equ.\ref{equ:5}.
\begin{equation}\label{equ:6}
\begin{aligned}
&\mathop{\min}_{\theta,\xi}d(\mathcal{G}_{\xi}(\mathcal{N}_{\theta}(x_{i}),f_{j};\{x_{i},x_{j}\}))\\
&\begin{array}{ll}
s.t. &f_{j}\triangleq\mathcal{N}_{\theta}(x_{j})
\end{array}
\end{aligned}
\end{equation}
For the $\mathcal{G}_{\xi}$, it is a learnable metric which is embedded the prior of topological invariance. It embeds the two original 3D medical images $x_{i},x_{j}$ which have consistent topology (Introduction section) into the calculation of the distance, and models the measurement of the distance for two features $f_{i}\triangleq\mathcal{N}_{\theta}(x_{i}),f_{j}$ as the measurement of the alignment degree for two image $x_{i},x_{j}$. Therefore, as shown in FIg.\ref{supp:ours}, this implicitly projects features onto a manifold of consistent topology (the invariant distribution of semantic regions in 3D medical images $\{x_{i},x_{j}\}$), and gathers the semantic features ($\mathcal{N}_{\theta}(x^{s1}_{i}),\mathcal{N}_{\theta}(x^{s1}_{j})$, $s1$ means the semantic regions on images) which are closed on this manifold.

\begin{algorithm*}[htb]
    \caption{GVSL: \textbf{G}eometric \textbf{V}isual \textbf{S}imilarity \textbf{L}earning}
    \label{alg:GVSL}
    \KwIn{

     $\mathcal{D},\mathcal{T}$\ \ \ \ \ \ \ \ \ \ \ dataset and the distribution of transformations;

     $\theta,\mathcal{N}_{\theta}$\ \ \ \ \ \ \ \ \ \ \ initial parameters for backbone network, backbone network;

     $\xi,\mathcal{G}_{\xi}$\ \ \ \ \ \ \ \ \ \ \ \ initial parameters for GM, GM head;

     $\iota,\mathcal{R}_{\iota}$\ \ \ \ \ \ \ \ \ \ \ \ initial parameters for self-restoration, restoration head;

     $optimizer$ \ optimizer, updates parameters via gradient;

     $K,N,\eta$\ \ \ \ \ \ \ iteration number, batch size, and learning rate.
     }
	\BlankLine

    \For{$k=1...K$}
        {$\mathcal{B}\leftarrow\{\{x^{i}_{A},x^{i}_{B}\}\sim\mathcal{D}\}^{N}_{i=1}$\tcp*{sample two batches from dataset}

        \For{$i,\{x_{A},x_{B}\}\in\mathcal{B}$}{

            $t\sim\mathcal{T}$\tcp*{sample image transformation}

            $x_{A}^{t}\leftarrow t(x_{A})$\tcp*{transform image $x_{A}$}

            $\{f_{A}^{g},f_{A}^{l}\}\leftarrow N_{\theta}(x_{A}^{t})$ and $\{f_{B}^{g},f_{B}^{l}\}\leftarrow N_{\theta}(x_{B})$\tcp*{compute global and local features from two images}

            $x_{A}^{'}\leftarrow \mathcal{R}_{\iota}(f_{A}^{l})$\tcp*{restore image $x_{A}$ in restoration head}

            $\psi_{AB}\leftarrow\mathcal{G}_{\xi}(f_{A}^{l},f_{B}^{g},f_{A}^{g},f_{B}^{g})$\tcp*{estimate a displacement vector field}

            $x_{AB}\leftarrow \psi_{AB}(x_{A})$\tcp*{align image $x_{A}$ to $x_{B}$}

            $l^{GVSL,i}_{\theta,\xi}\leftarrow l^{NCC}_{\theta,\xi}(x_{AB},x_{B})+l^{smooth}_{\theta,\xi}(\psi_{AB})$\tcp*{calculate the NCC loss and smooth loss for GVSL}

            $l^{MSE,i}_{\theta,\iota}\leftarrow \|x_{A}^{'}-x_{A}\|^{2}$\tcp*{calculate the MSE loss for self-restoration}}

            $\delta\theta\leftarrow\frac{1}{N}\sum_{i=1}^{N}(\partial_{\theta}l^{GVSL,i}_{\theta,\xi}+l^{MSE,i}_{\theta,\iota})$;

            $\delta\xi\leftarrow\frac{1}{N}\sum_{i=1}^{N}\partial_{\xi}l^{GVSL,i}_{\theta,\xi}$;

            $\delta\iota\leftarrow\frac{1}{N}\sum_{i=1}^{N}\partial_{\iota}l^{MSE,i}_{\theta,\iota}$\tcp*{compute the loss gradient w.r.t. $\theta$, $\xi$, and $\iota$}

            $\theta\leftarrow optimizer(\theta,\delta\theta,\eta)$;

            $\xi\leftarrow optimizer(\xi,\delta\xi,\eta)$;

            $\iota\leftarrow optimizer(\iota,\delta\iota,\eta)$\tcp*{update parameters i.e. $\theta$, $\xi$, and $\iota$}
            }

	\KwOut{$\mathcal{N}_{\theta}$\tcp*{the pre-trained backbone networks}}

\end{algorithm*}

\section*{Appendix B. Algorithm}
As illustrated in Alg.\ref{alg:GVSL}, our GVSL framework learns the GM between two images and the self-restoration as a baseline for consistent representation of same semantics between images.
\section*{Appendix C. Details of Transformation Operation $\mathcal{T}$}
\begin{figure}[htb]
  \centering
  \includegraphics[width=\linewidth]{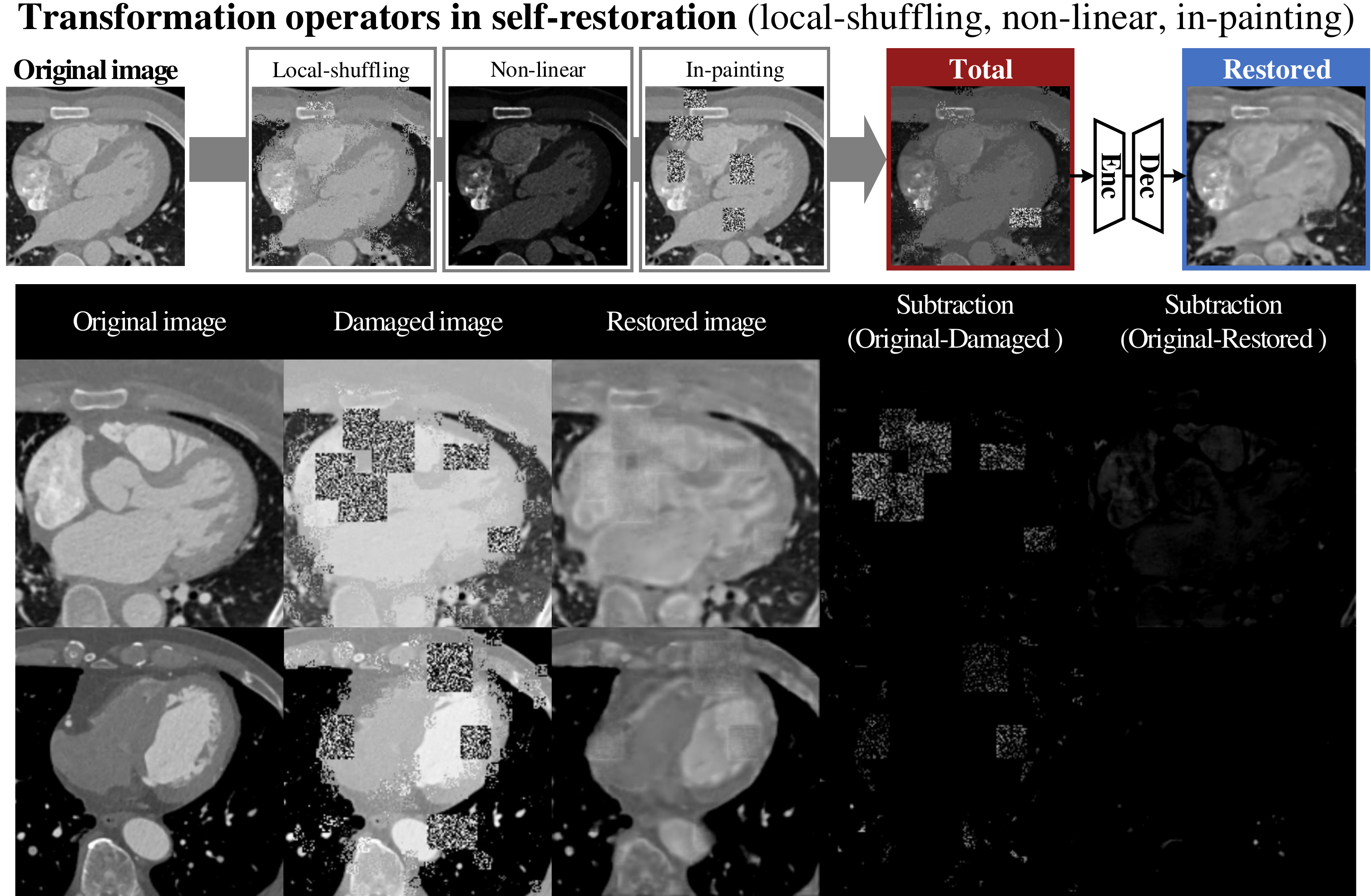}
  \caption{The visualization of the transformation operations. We utilize the in-painting, local-shuffling, and non-linear to construct the transformation distribution $\mathcal{T}$.}\label{supp:trans}
\end{figure}

During self-supervised training, our GVSL uses the consistency of following image transformation operations:
\begin{itemize}
  \item Random in-painting: This operation randomly selects 3D boxes inner images and the contents of these regions are replaced by the noise from a uniform distribution. Therefore, when learning the self-restoration and our GVSL, the network $\mathcal{N}_{\theta}$ will learn the dependency between the semantics and their context.
  \item Random local-shuffling: This operation randomly selects 3D boxes inner images and shuffles the voxels in the box regions. Therefore, when learning the self-restoration and our GVSL, the network $\mathcal{N}_{\theta}$ will learn the representation of texture features for semantics.
  \item Random non-linear transformation: This operation uses B\'{e}zier Curve which assigns every voxel a unique value via transform the distribution function of image. Therefore, the network $\mathcal{N}_{\theta}$ will learn the intensity information of semantic regions during the learning of self-restoration and our GVSL.
\end{itemize}
More specific related introductions can be find in the paper \cite{zhou2019models} which we follows. The Fig.\ref{supp:trans} demonstrates the transformation operations visually.

\section*{Appendix D. Details in our GVSL}
\begin{figure*}
\centering
\includegraphics[width=\linewidth]{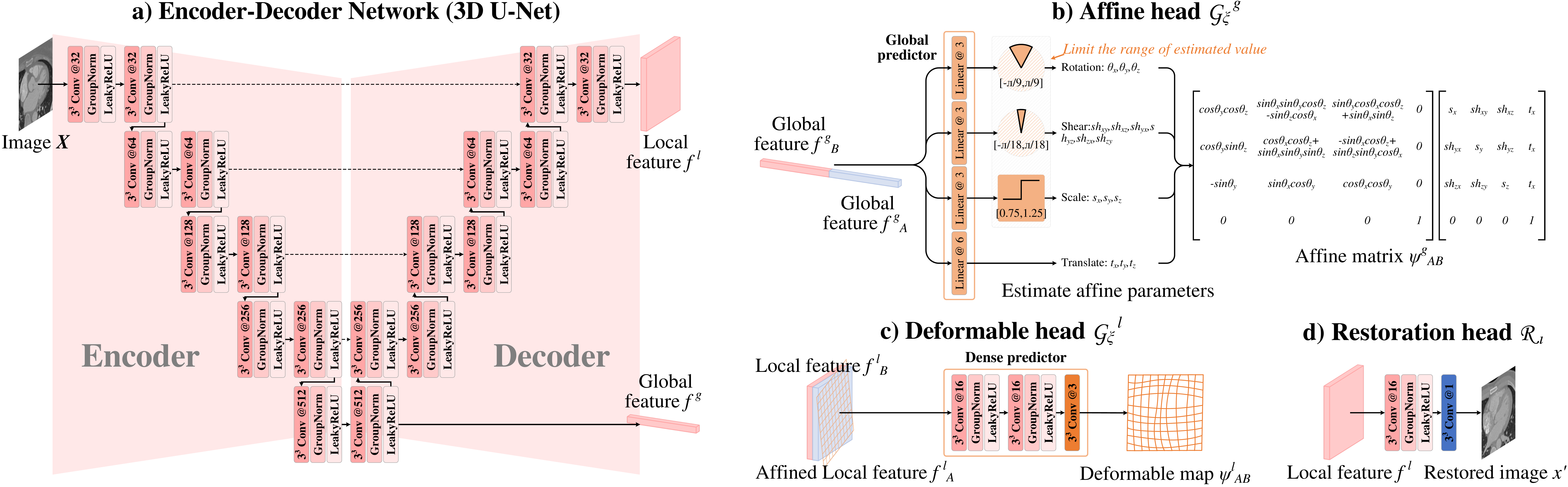}
\caption{\textbf{The details in our framework.} a) We take the 3D U-Net as our backbone, the features from the bottleneck and the final layer are the global features $f^{g}$ and the local features $f^{l}$. b) Affine head utilizes four linear layers to estimate affine parameters of rotation, shear, scale and translation. These parameters are used to make an affine matrix $\psi^{'}$ for affine transformation. c) Deformable head takes two Conv-groups followed by a convolution to estimate the deformable map $\phi^{'}$ via the local features. d) The restoration head takes a Conv-group followed by a convolution to restore the image.}
\label{supp:frameworkdetail}
\end{figure*}
\subsubsection*{D.1. Details in spatial transformation} We utilize the spatial transformation following \cite{jaderberg2015spatial} which is the function \emph{torch.nn.functional.grid\_sample} in PyTorch. For each voxel $p$ in image $x$, the DVF $\psi$ displaces the $p$ to a new (subpixel) voxel location $\psi(x(p))$ in image space. Then, the voxel in subpixel position is linearly interpolated to a near integer location at eight neighboring voxels. This process is formulated as
\begin{equation}
\label{equ:stn}
\psi(x(p))=\Sigma_{q\in\psi(\mathbb{Z}(p))}x(q)\Pi_{d\in\{x,y,z\}}(1-|\psi_{d}(x(p))-q_{d}|),
\end{equation}
where $\psi(\mathbb{Z}(p))$ are the voxel neighbors of $\psi(x(p))$, $\{x,y,z\}$ are the $x,y,z$ axes of 3D image.
\subsubsection*{D.2. Details in the network $\mathcal{N}_{\theta}$}
We utilize the 3D U-Net \cite{cciccek20163d} which is widely used in 3D medical images as the backbone network $\mathcal{N}_{\theta}$ in our framework. Owing to the limitation of GPU memory, we only use the batch size of 1 in our transferring process, and the batch size of 2 in our pre-training process. To avoid the overfitting problem caused by the Batch Normalization (BN) \cite{ioffe2015batch}, we utilize the Group Normalization \cite{wu2018group} to replace the BN in the original network.
\subsubsection*{D.3. Details in the fusion operation for DVF $\odot$}
As demonstrated in Equ.\ref{equ:def}, the affine matrix \cite{andrei20063d} utilizes the matrix consists of the rotation matrix, scaling matrix, shearing matrix, and translation matrix to make a movement for each voxels, thus achieving a global spatial transformation. This affine matrix $\psi^{g}_{AB}$ multiplies the position index $p=\{p_x,p_y,p_z\}$ of the voxel in image grid for the affine transformed position index $\hat{p}=\{p_{x},p_{y},p_{z}\}$. The transformed position index $\hat{p}$ is subtracted to the original position index $p$ for the affine vector and the affine vector is further added to the deformation vector in the position index $\hat{p}$ of deformation field $\psi^{l}_{AB}$ (Equ.\ref{equ:def2}), thus achieving the vector to move the voxel in position $\psi_{AB}(p)$. This operation is performed for whole positions in the image grid, fusing the affine matrix and the deformation field for the DVF $\psi_{AB}$.
\begin{figure*}
  \centering
\begin{align}
&\begin{array}{c}\label{equ:def}
  \psi^{g}_{AB}=
 \overbrace{\begin{bmatrix}
  \cos\theta_{y}\cos\theta_{z} & \sin\theta_{x}\sin\theta_{y}\cos\theta_{z}-\sin\theta_{z}\cos\theta_{x} & \sin\theta_{y}\cos\theta_{x}\cos\theta_{z}+\sin\theta_{x}\cos\theta_{z} & 0 \\
  \cos\theta_{y}\sin\theta_{z} & \cos\theta_{x}\cos\theta_{z}+\sin\theta_{x}\sin\theta_{y}\sin\theta_{z} & -\sin\theta_{x}\cos\theta_{z}+\sin\theta_{z}\sin\theta_{y}\cos\theta_{x} & 0 \\
  -\sin\theta_{y} & \sin\theta_{x}\cos\theta_{y} & \cos\theta_{x}\cos\theta_{y} & 0 \\
  0 & 0 & 0 & 1
\end{bmatrix}}^{\textbf{\text{Rotation}}}\\
\underbrace{\begin{bmatrix}
s_{x} & 0 & 0 & 0 \\
0 & s_{y} & 0 & 0 \\
0 & 0 & s_{z} & 0 \\
0 & 0 & 0 & 1
\end{bmatrix}}_{\textbf{\text{Scaling}}}
\underbrace{\begin{bmatrix}
1 & sh_{yx} & sh_{zx} & 0 \\
sh_{xy} & 1 & sh_{zy} & 0 \\
sh_{xz} & sh_{yz} & 1 & 0 \\
0 & 0 & 0 & 1
\end{bmatrix}}_{\textbf{\text{Shearing}}}
\underbrace{\begin{bmatrix}
1 & 0 & 0 & t_{x} \\
0 & 1 & 0 & t_{y} \\
0 & 0 & 1 & t_{z} \\
0 & 0 & 0 & 1
\end{bmatrix}}_{\textbf{\text{Translation}}},
\end{array}\\
&\begin{array}{c}\label{equ:def2}
\psi_{AB}(p)\triangleq\psi^{l}_{AB}(p)\odot\psi^{g}_{AB}=\begin{bmatrix}
\psi_{AB}(p_{x}) \\
\psi_{AB}(p_{y}) \\
\psi_{AB}(p_{z}) \\
\end{bmatrix}=
\begin{bmatrix}
\psi^{l}_{AB}(\hat{p_{x}}) \\
\psi^{l}_{AB}(\hat{p_{y}}) \\
\psi^{l}_{AB}(\hat{p_{z}}) \\
\end{bmatrix}+
\begin{bmatrix}
\hat{p_x} \\
\hat{p_y} \\
\hat{p_z} \\
\end{bmatrix}-
\begin{bmatrix}
p_x \\
p_y \\
p_z \\
\end{bmatrix},
\begin{bmatrix}
\hat{p_{x}} \\
\hat{p_{y}} \\
\hat{p_{z}} \\
1
\end{bmatrix}=
\psi^{g}_{AB}\times\begin{bmatrix}
p_{x} \\
p_{y} \\
p_{z} \\
1
\end{bmatrix}
\end{array}
\end{align}
\end{figure*}
\section*{Appendix E. Details of Datasets and Implementations in Experiment}
\begin{table*}[h]
\centering
\caption{The details of the clinical dataset in our pretext task and four public available datasets in our downstream tasks.}
\resizebox{\textwidth}{!}{
\begin{threeparttable}
\begin{tabular}{|c|c|c|c|l|}
\hline
\multicolumn{5}{|c|}{\textbf{a)} The details of four public available datasets in \textbf{downstream tasks}}\\
\hline
\textbf{Name}   &\textbf{Target dataset}    &\textbf{Train/Val/Test}    &\textbf{Downstream task}&\textbf{Pre-processing}\\
\hline
SHC          &MM-WHS 2017 CT$^{a}$       &15/5/40                    &Segmentation of 7 heart structures&
\begin{tabular}[c]{@{}l@{}}
1.Crop the heart regions\\
2.Resample the resolution to $1mm^{3}$\\
3.Normailze via $\frac{\max(\min(0,x),2048)}{2048}$\end{tabular}\\
\hline
SAC          &ASOCA 2020 CT$^{b}$        &15/5/20                    &Segmentation of coronary artery&
\begin{tabular}[c]{@{}l@{}}
1.Crop heart regions\\
2.Resample the resolution to $1mm^{3}$\\
3.Normailze via $\frac{\max(\min(0,x),2048)}{2048}$\end{tabular}\\
\hline
SBM          &CANDI MR$^{c}$             &40/20/43                   &Segmentation of 28 brain tissues&
\begin{tabular}[c]{@{}l@{}}
1.Crop $160^{2}\times128$ regions around brain\\
2.Resample the resolution to $1mm^{3}$\\
3.Normailze via $\frac{x-\min(x)}{\max(x)-\min(x)}$\end{tabular}\\
\hline
CCC          &STOIC CT$^{d}$             &1000/400/600               &Diagnosis of COVID-19&
\begin{tabular}[c]{@{}l@{}}
1.Extract lung regions via lungmask$^{e}$\\
2.Resample the resolution to $1mm^{3}$\\
3.Normailze via $\frac{\max(\min(0,x),2048)}{2048}$\end{tabular}\\
\hline
\multicolumn{5}{|c|}{\textbf{b)} The details of the clinical dataset in \textbf{pretext task}}\\
\hline
\textbf{Amount} &\textbf{Image type}        &\multicolumn{2}{c|}{\textbf{Detail information}}     &\textbf{Pre-processing}\\
\hline
302             &Coronary CT angiography    &\multicolumn{2}{l|}{\begin{tabular}[c]{@{}l@{}}
1.Scanner: SOMATOM Definition Flash\\
2.x/y-resolution: 0.25$\sim$0.57 mm/voxel\\
3.Slice thickness: 0.75$\sim$3 mm/voxel\\
4.x/y-size: 512 voxels, z-size: 128$\sim$994 voxels\end{tabular}}&
\begin{tabular}[c]{@{}l@{}}
1.Resample the resolution to $1mm^{3}$\\
2.Normailze via $\frac{\max(\min(0,x),2048)}{2048}$\end{tabular}\\
\hline
\end{tabular}
\begin{tablenotes}
     \item[a] MM-WHS 2017: \url{http://www.sdspeople.fudan.edu.cn/zhuangxiahai/0/mmwhs/}
     \item[b] ASOCA: \url{https://asoca.grand-challenge.org/}
     \item[c] CANDI: \url{https://www.nitrc.org/projects/candi\_share/}
     \item[d] STOIC challenge: \url{https://stoic2021.grand-challenge.org/stoic-db/}
     \item[e] Lungmask code: \url{https://github.com/JoHof/lungmask}
\end{tablenotes}
\end{threeparttable}}
\label{tab:1}
\end{table*}
As shown in Tab.\ref{tab:1}, we pre-train the network on the pre-training dataset and evaluate the models on four downstream tasks with different properties, giving a complete evaluation.
\subsection*{E.1. Details of the pre-training dataset}
The pre-training dataset consists of 302 cardiac CT images with numerous semantic regions. These images were scanned on a SOMATOM Definition Flash and the contrast media was injected during the scanning process. The x/y-resolution of these CT images is between 0.25 to 0.57 mm/voxel and the slice thickness is between 0.75 to 3 mm/voxel. The x/y-size of the images is 512 voxels and the z-size is between 128  to 994 voxels. For pre-processing, we firstly resample the resolution of these images to $1mm\times1mm\times1mm$ for a unified resolution, then threshold their grayscale value to [0, 2048] and normalize them to [0, 1] for unified intensity.
\subsection*{E.2. Details of the downstream datasets}
\paragraph{E.2.1. The SHC task} targets on segmenting seven large heart structures on the CT images from MM-WHS 2017 dataset \cite{zhuang2019evaluation} which originally has 20 image-label pairs and 40 unlabeled images. We randomly split 15 of the image-label pairs as the training set, 5 of them as the validation set, and the original 40 unlabeled images as the testing set and test the results on the officially provided software. For pre-processing, we firstly crop the heart regions of interests (ROIs) to reduce the size due to the limited GPU memory and resample the resolution of these images to $1mm\times1mm\times1mm$ for a unified resolution. These images are further thresholded to [0, 2048] grayscale value, and normalized to [0, 1] via dividing by 2048 for unified intensity. This task evaluate the \textbf{inner}-scene transferring ability of the models on a \textbf{dense} prediction task for \textbf{large} structures.

\paragraph{E.2.2. The SAC task} targets on segmenting the small coronary arteries on the Coronary CT Angiography (CCTA) images from ASOCA dataset \cite{gharleghi2022automated} which originally has 40 image-label pairs. We randomly split 15 of them as the training set, 5 of them as the validation set, and 20 of them as the testing set. Following the SHC task, we also crop the heart ROIs, resample their resolution to $1mm\times1mm\times1mm$, threshold the grayscale to [0, 2048] and normalize the intensity to [0, 1] via dividing by 2048. This task evaluate the \textbf{inner}-scene transferring ability of the models on a \textbf{dense} prediction task for \textbf{small} structures.

\paragraph{E.2.3. The SBM task} targets on segmenting 28 brain tissues on the brain T1-weighted MR images from CANDI dataset \cite{kennedy2012candishare} which has 103 image-label pairs. We randomly split 40 of them as the training set, 20 of them as the validation set, and 43 of them as the testing set. Following some works \cite{Wang2020CVPR,ding2021modeling} on this dataset, we crop a $160\times160\times128$ region around the center of the brain which contain the whole brain for computation efficiency. The grayscale value of these images are further limited bigger than 0, and normalized to [0, 1] via min-max normalization for unified intensity. This task evaluate the \textbf{inter}-scene transferring ability of the models on a \textbf{dense} prediction task for \textbf{multiple} (28) structures.

\paragraph{E.2.4. The CCC task} targets on classifying (diagnosis) the COVID-19 or the health on chest CT images from the STOIC challenge dataset \cite{revel2021study} which originally has 2000 public training set. To evaluate the models in our experiment, we further randomly split 1000 of them as the training set, 400 of them as the validation set, and 600 of them as the testing set. For pre-processing, we extract the lung regions via the existing open released code of lungmask to remove the interruption of the background, crop the lung ROIs to reduce the size, and resample the resolution of these images to $1mm\times1mm\times1mm$ for a unified resolution. Following the H.CT.S task, we also threshold the grayscale to [0, 2048] and normalize the intensity to [0, 1] via dividing by 2048. This task evaluate the \textbf{inner}-scene transferring ability of the models on a \textbf{global} prediction task.

\subsection*{E.3. Implementation details of transfer learning on downstream tasks}
\paragraph{E.3.1. Implementation for linear evaluation}
We take linear evaluation to evaluate the clustering effect of the extracted features thus demonstrating the representability of the pre-trained network. 1) For segmentation tasks (SHC, SAC, SBM), we use the whole pre-trained backbone network $\mathcal{N}_{\theta}$ as a fixed feature extractor for the new downstream datasets. And then, the local features $f^{l}$ from the decoder of the network are used to train a convolutional layer followed with a Softmax activation function. 2) Like the implementation for segmentation tasks, for classification task (CCC), we use the encoder part of the backbone network $\mathcal{N}_{\theta}$ as a fixed feature extractor for downstream tasks. And then, the global features $f^{g}$ from the fixed encoder are used to train a linear layer followed with a Sigmoid activation function in CCC task for the evaluation of the representability for global features. We use a batch size of 1 due to the limitation of GPU memory and a learning rate of $1\times10^{-4}$ with Adam \cite{kingma2014adam} optimizer to train these tasks, and save the parameters with the highest DSC or AUC score on validation sets for segmentation or classification tasks.
\paragraph{E.3.2. Implementation for fine-tuning evaluation}
We further take fine-tuning evaluation to evaluate the transferring ability thus demonstrating the great potential for initialization of downstream tasks. We most follow Models Genesis \cite{zhou2019models} for training fine-tuning models. 1) For segmentation tasks (SHC, SAC, SBM), we connect the whole backbone network $\mathcal{N}_{\theta}$ with a convolutional layer followed by a Softmax activation function, thus constructing a segmentation framework. The gradient optimizes the all parameters in this framework during the backpropagation. 2) For classification task (CCC), we use the encoder part of the backbone network, and the encoder is connected to a linear layer followed with a Sigmoid activation function. Like the segmentation tasks, the gradient optimizes all parameters in the framework. Like the implementation of linear evaluation, we also use the batch size of 1 and learning rate of $1\times10^{-4}$ with Adam \cite{kingma2014adam} optimizer, and save the parameters with the highest DSC or AUC score on validation sets.

\end{document}